%% file: icml_2021_paper.tex
\documentclass{article}

\usepackage{microtype}
\usepackage{graphicx}
\usepackage{subfigure}
\usepackage{booktabs} 

\usepackage{color-edits}

\addauthor{mp}{red}
\addauthor{tc}{blue}
\addauthor{vf}{green}
\addauthor{et}{magenta}

\usepackage{amsfonts}
\usepackage{amsmath}
\usepackage{amssymb}
\usepackage{amsthm}
\usepackage{nicefrac}
\usepackage{algorithm}
\usepackage{algpseudocode}
\usepackage{hyperref}

\newtheorem{lemma}{Lemma}
\newtheorem{theorem}{Theorem}

\newcommand{\sdef}{\overset {\mathrm {def}} {=}}
\newcommand{\ev}[2]{\underset {#1} {\mathbb E} \left [ #2 \right ]}
\newcommand{\var}[2]{\underset {#1} {\mathrm {Covar}} \left [ #2 \right ]}
\newcommand{\pr}[2]{\Pr_{#1} \left [ #2 \right ]}
\renewcommand{\hat}[1]{\widehat {#1}}
\newcommand{\nrv}[2]{\mathcal {N} \left ( #1, #2 \right )}

\newcommand{\pref}[1]{\ref{#1}}

\newcommand{\rb}[1]{\rotatebox{0}{#1}}

\newcommand{\eigen}{\href{http://eigen.tuxfamily.org}{\textcolor{blue}{\texttt{eigen}}}}

\newcommand{\stan}{\href{https://mc-stan.org/}{\textcolor{blue}{\texttt{stan}}}}
\newcommand{\tensorflow}{\href{https://tensorflow.org/}{\textcolor{blue}{\texttt{TensorFlow}}}}
\newcommand{\pyro}{\href{https://pyro.ai/}{\textcolor{blue}{\texttt{pyro}}}}

\usepackage{hyperref}

\usepackage{hyperref}

\usepackage[accepted]{icml2021}

\icmltitlerunning{Truncated Log-concave Sampling with Reflective Hamiltonian Monte Carlo}

\begin{document}

\twocolumn[
\icmltitle{Truncated Log-concave Sampling with Reflective Hamiltonian Monte Carlo}




\begin{icmlauthorlist}
\icmlauthor{Apostolos Chalkis}{nkua}
\icmlauthor{Vissarion Fisikopoulos}{nkua}
\icmlauthor{Marios Papachristou}{cornell}
\icmlauthor{Elias Tsigaridas}{inria}

\end{icmlauthorlist}

\icmlaffiliation{nkua}{National Kapodistrian University of Athens}
\icmlaffiliation{cornell}{Cornell University}
\icmlaffiliation{inria}{INRIA Paris}

\icmlcorrespondingauthor{Marios Papachristou}{papachristoumarios@cs.cornell.edu}

\icmlkeywords{Machine Learning, ICML}

\vskip 0.3in
]



\printAffiliationsAndNotice{Author names in alphabetical order.} 

\begin{abstract}
We introduce Reflective Hamiltonian Monte Carlo (ReHMC),
an HMC-based algorithm, to sample from a log-concave distribution restricted to a convex polytope.
We prove that, starting from a warm start, it mixes in
$\widetilde O(\kappa d^2 \ell^2 \log (1 / \varepsilon))$ 
steps for a well-rounded polytope, ignoring logarithmic factors
where $\kappa$ is the condition number of the negative log-density, $d$ is the dimension, 
$\ell$ is an upper bound on the number of reflections, and $\varepsilon$ is the accuracy parameter. We also developed an open source implementation of ReHMC and we performed an experimental study on various high-dimensional data-sets. Experiments suggest that ReHMC outperfroms Hit-and-Run and Coordinate-Hit-and-Run regarding the time it needs to produce an independent sample.
\end{abstract}

\section{Introduction}

One particularly interesting and fundamental computational problem is sampling from a high-dimensional log-concave density of the form $\pi(x) \propto e^{-f(x)}$ constrained in a convex polytope $K \subset \mathbb R^d$, where $f$ is an $L$-smooth and $m$-strongly-convex function, with a Markov Chain Monte Carlo (MCMC) method. This problem appears commonly in machine learning \cite{brock2018large}, finance \cite{cales2018practical}, numerical analysis \cite{cousins2015bypassing}, optimal control \cite{ huynh2012incremental, he2017numerical}, Bayesian inference \cite{gamerman2006markov}, computational geometry \cite{dyer1991random}, and in many more areas. Example problems consist of the Bayesian logistic regression, Bayesian mixture models, Gaussian sampling, and flux sampling from metabolic networks \cite{Herrmann19}. 

Current research on MCMC methods focuses on both practical and theoretical aspects and there are efficient implementations, like \tensorflow~\cite{abadi2016tensorflow}, \stan~\cite{carpenter2017stan}, and \pyro~\cite{bingham2019pyro}, that provide MCMC methods for sampling,
that in turn allow the creation of powerful Bayesian models. 
Perhaps one of the best-known algorithms for sampling from a log-concave density $e^{-f(x)}$ is the first-order method of Hamiltonian Monte Carlo (HMC). HMC simulates an imaginary particle moving in a conservative field determined by a negative log-probability function $f(x)$ and its gradient $\nabla f(x)$. In this setting, we bound the number iterations that HMC performs to converge to the target distribution $\pi$, when we restrict it to a convex polytope $K$. In addition we develop a scalable (up to thousands of dimensions) implementation that outperforms other contemporary methods. 

Our work extends a series of results~\cite{lee2020logsmooth, chen2019fast, dwivedi2019log} which devise mixing time guarantees for first-order methods. Even though our strategy finds its roots in these results, our analysis is different to account for convex-body domains, in our case polytopes, since the Hamiltonian dynamics exhibit \emph{reflections} at the boundaries. The main obstacle to overcome is that when HMC is done with reflections, the proposal distributions for the new points are not purely Gaussian; actually they are mixtures of Gaussians. This \emph{differentiates} the constrained from the unconstrained case and demands new techniques
to bound the number of iterations.


\textbf{Our Contribution.} We introduce Reflective Hamiltonian Monte Carlo (ReHMC), a HMC-based algorithm to sample from a truncated log-concave density using leapfrog dynamics with boundary reflections in a polytope. Under mild assumptions, we prove that ReHMC gives a sample that is $\varepsilon$-close, in terms of the total variation distance,  to the target density $\pi$ in $O(\kappa d^2 \ell^2 \log^2(\kappa / \varepsilon) \log (d \log(\kappa /  \varepsilon) + d \log (\gamma / \varepsilon)) \log (1 / \varepsilon))$ iterations, where $\kappa$  is the condition number of $f$, $d$ is the dimension, $\ell$ is a (high-probability) upper bound on the maximum number of reflections occurring in the body,
and $\gamma$ is the sandwiching ratio of the convex polytope.  Our bound has a linear dependence on the condition number $\kappa$, in agreement to the current state-of-the-art, a doubly logarithmic dependence on $\gamma^d$, an improvement compared to the (poly) $\gamma^2$-dependence that Hit-and-Run (H\&R) has on the geometry of the body~\cite{lovasz2006hit}, and a quadratic dependence on $d$, similar to H\&R. 
Furthermore, this is the first analysis of an HMC-based algorithm that exploits reflections. In this random walk, the proposal and transition distributions are different from the unconstrained case of~\cite{lee2020logsmooth, chen2019fast, dwivedi2019log} and thus we cannot reuse their tools and techniques in our analysis. 
As a corollary of our analysis we get a mixing time result of $O(d^2 \ell^2 \log(d \log(\gamma / \varepsilon)) \log^3 (1 / \varepsilon)$ for sampling from a uniform density with a billiard-based walk similar to~\cite{gryazina2014random} which has an \emph{unknown} mixing time. 

Regarding the practical nature of ReHMC, we developed an open-source high-performance implementation in C++ that scales to thousands of dimensions.
We compare ReHMC with the H\&R
algorithm \cite{smith1996hit, lovasz2006hit, shen2020composite} that is commonly used in  modern toolboxes, e.g., COBRA \cite{becker2007quantitative} and HOPS \cite{jadebeck2020hops}, on a wide variety of convex polytopes.
In particular, we perform various experiments by sampling 
from well-known polytopes, like cubes, simplices, products of simplices, Birkhoff polytopes and cross polytopes,
as well as from polytopes coming from structural biology;
for the latter efficient sampling from a truncated log-concave density is of crucial importance~\cite{Herrmann19, cousins2017efficient}. 
We evaluate the practical performance by the rate that a mixed Markov Chain produces \emph{independent samples} \cite{geyer2011introduction} using the PSRF diagnostic \cite{gelman1992inference}.
ReHMC scales up to $\sim 10^3 \times$ faster than H\&R when we sample from a unit-covariance Gaussian density centered at the Chebyshev center of the polytope \cite{Boyd04}.

\textbf{Code Availability:}~\cite{anonymous_2021_4459362}.

\textbf{Notation.} We sample from a distribution $\Pi$ with PDF $\pi(x) \propto e^{-f(x)}$, where  $f(x)$ is a convex function with support on $K \sdef \mathrm {supp}(\Pi) = \{ x \in \mathbb R^d \mid \pi(x) > 0 \} \subset \mathbb R^{d}$ and $K$ is a convex polytope with an interior $K^o$ and boundary $\partial K$. The function $f$ has a minimizer at $x^* \in K^o$.
The sandwiching ratio of $K$ is  $\gamma \sdef \inf_{R > r > 0} \left \{ R / r \mid \mathbb B(x^*, r) \subseteq K \subseteq \mathbb B(x^*, R) \right \}$, where $\mathbb B(x^*, r)$ is the $d$-dimensional $L_2$ ball with radius $r$ centered at $x^*$. 
We assume that $f$ is twice differentiable in $K$, including the boundary $\partial K$, $L$-smooth and $m$-strongly convex. 
The Hessian of $f$ is $\nabla^2 f$ and it has eigenvalues in the range $[m, L]$ and condition number $\kappa \sdef L / m$. 
We use  $\Pi(A)$ to denote the measure of set $A$ under the distribution $\Pi$ whose its density is $\pi$.  
For two probability distributions $\Pi$ and $P$ on the same domain $K$ we define the \emph{total variation distance} (TVD) between them as $
    \| \Pi - P \|_{TV} \sdef \sup_{A \in \mathcal B(K)} | \Pi(A) - P(A) | = \frac 1 2 \int_{K} \left | \frac {d P (x)} {d \Pi(x)} - 1 \right | d \Pi(x) = \frac 1 2 \int_K | \pi(x) - \rho(x) | d x
$, where $\mathcal B(K)$ denotes the Borel $\sigma$-algebra of $K$.
Similarly the Kullback–Leibler (KL) Divergence is $
    d_{KL} (\Pi, P) = \int_K  \log  ( \pi(x) / \rho(x) ) \pi(x) dx
$.
The two statistical distances, TVD and KL, are connected via Pinkser's Inequality \cite{csiszar2011information}. 
For a function $h: K \to \mathbb R$ with respect to a distribution $\Pi$ with density $\pi$, we denote the expected value and the covariance by 
$\ev {\pi} {h(x)}$ and 
$\var {\pi} {h(x)} = \ev {\pi} {(h(x) - \ev {\pi} {h(x)}) (h(x) - \ev {\pi} {h(x)})^\top}$, respectively. We say that $P$ is $\beta$-warm with respect to $\Pi$ if and only if $\sup_{x \in K} dP / d \Pi = \sup_{x \in K} \rho(x) / \pi(x) \le \beta$.

For MCMC algorithms, we denote by $\mathcal P_x$ the proposal distribution given that the sampler is positioned at $x$; by $\mathcal T_x$ we denote the corresponding transition distribution given that the sampler is at $x$, where $\mathcal T$ denotes the transition operator. The \emph{ergodic flow} of a set $S \subseteq K$ is defined as $\mathcal Q(S) \sdef \int_S \mathcal T_x(S^c) \pi(x) dx$ where $S^c \sdef K \setminus S$. 
The distribution after $k$ steps is  $\pi_k \sdef \mathcal T^k \pi_0$ and we denote the \emph{average density at step $k$} \cite{kannan2006blocking} by $\nu_k(x) \sdef \frac 1 k \sum_{i \le k} \pi_i(x)$ with CDF $N_k$.

\section{Related Work}

\textbf{First-order Unconstrained Methods.} General first-order methods for sampling assume access to the density and its gradient and include some well-known methods for sampling; Underdamped Langrevin Dynamics (ULD)~\cite{lee2018algorithmic}, Metropolis-Adjusted Langevin Algorithm (MALA)~\cite{dwivedi2019log} and HMC~\cite{lee2020logsmooth, dang2019hamiltonian} are among the most famous ones. The current work on these methods assumes distributions  supported on $\mathbb R^d$, i.e., they do not pose any constraints on the domain of the samples. 
The recent bound for the mixing time of MALA 
is  $\tilde O(\max \{ \kappa d, \kappa^{1.5} \sqrt d \})$~\cite{dwivedi2019log}\footnote{The notation $\tilde O( \cdot )$ ignores polylogarithmic factors.}  which was improved to $\tilde O(\kappa d)$ in the work of~\cite{lee2020logsmooth}. Another recent work \cite{shen2019randomized}
proves convergence time bound of $\tilde O(\kappa^{7/6} / \varepsilon^{1/3} + \kappa / \varepsilon^{2/3}) $ in the 2-Wasserstein distance for  ULD using the randomized midpoint method.  

\textbf{Constrained HMC.} 
There are several works that have previously examined constrained versions of the HMC algorithm. More specifically, \cite{afshar2015reflection} examines the HMC variant with reflection and refraction using a Leapfrog integrator, but it does not analyze the \emph{mixing time} and the experimental part is restricted to low dimensions. 
The works of~\cite{pakman2015package, chevallier2020improved} proposed HMC methods with reflections combined with billiard trajectories where a billiard trajectory is planned and it is rejected if the total number of reflections exceeds a certain threshold.
This is a specialized approach for sampling from a Gaussian density with mean $\mu$ and covariance matrix $\Sigma$, where the Hamiltonian dynamics are relatively simple. \cite{chevallier2020improved}~give an $O(\log d)$ mixing time for cubes with an $O(d)$ number of reflections per-step on expectation
and also prove uniform ergodicity. Our work provides mixing time bounds for sampling general log-concave densities on general domains.
Both ours and the aforementioned algorithms are related to the theory of \emph{dynamical billiards}~\cite{masur2002rational, de2003geometric}.

Another important family of methods for HMC-based sampling considers the inclusion of the local geometry in the Hamiltonian via barrier functions (log-barriers and sigmoid barriers) \cite{girolami2009riemannian, betancourt2013general, yi2017roll, nishimura2016geometrically, lee2018convergence}; however the barrier functions become ill-conditioned near the boundaries \cite{wright1994some}. Finally, in this family of methods we should also include the work of \cite{shen2020composite} which samples from a density of the form $e^{-f(x) - g(x)}$, where $f$ and $g$ are convex, but $g$ is non-smooth \cite{mou2019efficient, shen2020composite, pereyra2016proximal, bubeck2018sampling, brosse2017sampling}. It achieves constrained sampling  using a non-smooth barrier function (such as the log-barrier). 

\textbf{Mixing of MCMC.} The rate of convergence of a Markov chain, i.e., its mixing time, is dependent on the conductance of the chain \cite{Jerrum88}. Roughly speaking, the conductance $\Phi$ of a Markov chain determines the ``maximum bottleneck" of the chain, i.e., it is the minimum ergodic flow subject to the target and the transition distribution between a subset $S$ of the space and its complement $S^c$ divided by the minimum measure of $S$ and $S^c$.  The classical result of~\cite{lovasz1990mixing} states that the mixing time is between $1 / \Phi$ and $\log \beta / \Phi^2$, where $\beta$ is the \emph{warmness} of the starting distribution. We can obtain improved mixing time bounds by refining this methodology, for instance via the average conductance method \cite{lovasz1999faster} and the blocking conductance framework of \cite{kannan2006blocking}. We use the blocking conductance framework and its refinement due to \cite{lee2020logsmooth} to obtain our convergence result. 

\textbf{Practical performance \& Software.} The main paradigm in practice is Coordinate Directions Hit-and-Run (CH\&R). Extended experiments~\cite{Emiris18, Cousins15} have shown that Hit-and-Run (H\&R) converges after $\tilde O(d^2)$ steps. CH\&R also converges after $\tilde O(d^2)$ steps in practice~\cite{Emiris18, haraldsdottir2017chrr} and moreover, when the truncation is given by a convex polytope, its cost per step is smaller than H\&R's cost per step. This is the main reason why CH\&R overshadowed, until recently, all other random walks in practical computations on polytopes.
Considering software for truncated sampling, COBRA~\cite{becker2007quantitative} provides CH\&R with a rounding preprocess for uniform and exponential sampling from convex polytopes that appear as flux spaces of metabolic networks. The package HOPS~\cite{jadebeck2020hops} provides both H\&R and CH\&R for general distribution combined with the same rounding preprocess before sampling as in COBRA. HOPS implementation of CH\&R outperforms COBRA as shown in~\cite{jadebeck2020hops}. 

\textbf{Truncated Statistics.} The study of truncated statistics gathers a lot of attention the recent years \cite{daskalakis2019computationally, ilyas2020theoretical, daskalakis2020truncated, o1995truncated} with a focus on many "classical problems", such that linear regression and logistic regression. In this setting, the phenomenon of \emph{output truncation} is studied, where the samples are filtered out wrt the values of the response variables. Truncation is  attributed to poor measurements and data collection as well as privacy concerns. These errors usually lead to biased models, i.e. models that replicate the biases of the data that they have been trained on. Lastly, truncation has also been studied through the lens of more advanced generative modeling through Generative Adversarial Networks~\cite{brock2018large, marchesi2017megapixel}.


\section{Algorithm} \label{sec:algorithm}


In the sequel we present the sampling algorithm ReHMC. First, we briefly introduce Hamiltonian dynamics and then their discretization using the leapfrog dynamics. Next,  we bound the mixing time of ReHMC and we highlight the key points of its analysis.

\subsection{ReHMC} 

\textbf{Hamiltonian Dynamics.
\footnote{For a more detailed introduction to the subject we redirect the interested reader to \cite{neal2011mcmc} and \cite{betancourt2017conceptual}.}} HMC simulates the movement of a particle to sample from a  distribution $\Pi$. The state of the particle consists of a position vector $x$ and momentum vector $v$ with a Hamiltonian function 
\begin{equation}
    \mathcal H(x, v) \sdef \underbrace{\tfrac 1 2 \| v \|^2}_{\text{Kinetic Energy}} + \underbrace{f(x)}_{\text{Potential Energy}} ,
\end{equation}
where $f$ is an $L$-smooth and $m$-strongly-convex function with condition number $\kappa$. The particle's movement evolves according to the Hamiltonian dynamics 
\begin{align} \label{eq:hmc}
    \frac {d x} {d t} = +\frac {\partial \mathcal H} {\partial v} =  + v, \;
    \frac {d v} {d t} = -\frac {\partial \mathcal H} {\partial x} = - \nabla f(x),
\end{align}
that ideally preserve the Hamiltonian $\mathcal H$. If the particle is restricted to a convex polytope $K$, then it faces an \emph{infinite potential barrier}. Hence, it \emph{reflects} at $\partial K$
and Hamiltonian dynamics embody the reflections \cite{gryazina2014random, afshar2015reflection}.The dynamics are volume preserving and time-reversible (App.~\ref{sec:volume_preservation_ideal}). In the MCMC regime, this setting allows us to sample from a Markov chain with joint stationary distribution proportional to
\begin{equation} \label{eq:stationary}
    e^{- \| v \|^2 / 2} \cdot \underbrace{e^{-f(x)} \mathbf 1 \{ x \in K \}}_{\text{Target density $\pi$}} ,
\end{equation}
where the $x$-marginal is the (truncated) target distribution $\pi$. The simulation of the \emph{continuous} Markov chain, assuming that the sampler is positioned at $x$, is as follows: First, we draw an initial velocity $v \sim \nrv 0 {I_d}$ and simulate the (reflective) Hamiltonian dynamics with initial conditions $x(0) = x$ and $v(0) = v$ for $s$ units of time. Then, the new state $(x(s), v(s))$ is proposed and 
we apply a Metropolis filter to preserve the stationary distribution of \eqref{eq:stationary}. That is we perform a coin flip with bias
\begin{equation} \label{eq:metropolis_filter}
    \min \left \{ 1, e^{-\mathcal H(x(s), v(s)) + \mathcal H(x(0), v(0))} \right \}.
\end{equation}
Finally, if the coin comes up heads, then the sampler
moves to the proposed state $(x(s), v(s))$; otherwise it remains  at $(x(0), v(0))$. 
In the case of the continuous dynamics, the Hamiltonian is \emph{exactly} preserved, that is $\dot {\mathcal H} = 0$ and the value of the filter equals 1, and the sampler always moves at the proposed position. For \emph{discretized} dynamics that we use in computer simulations, the value of the filter does not equal 1 and the proposed sample may be rejected. The pseudocode of the procedure appears in App.~\ref{sec:truncated_hmc} (Alg.~\ref{alg:truncated_hmc}). These dynamics are volume-preserving and time-reversible (App.~\ref{sec:reflective_dynamics_volume_preserving}).

\textbf{Oracle model.} The oracle model has access to $f$, $\nabla f$ 
and to a boundary oracle that computes the (intersection) point  $\partial K \cap \{ z \in \mathbb R^d | z = (1-t)x + ty, t \in [0,1],\ x\in K \}$. If $K$ has $M$ facets and its representation is $K = \{ x \,|\, Ax \le b,  \text{ where } A \in \mathbb R^{M \times d}, b \in \mathbb R^M, \text{and }\| a_i \| = 1 \}$, the intersection point is the smallest possible $t_i\in [0,1]$ such that $a_i^\top ((1-t_i)x + t_iy) = b_i$. Clearly, the computation of each $t_i,\ 1\le i\leq M$ takes $O(d)$ operations.

\textbf{Discretization.} We use the symplectic method of leapfrog integration to discretize the dynamics of \eqref{eq:hmc}. The discretized version of reflective Hamiltonian Dynamics  updates the initial state $(x, v)$ at time $t$
to the state $(\tilde x', \tilde v')$ at time $t + \eta$
with the leapfrog integrator
by \emph{velocity half-update} and the \emph{position update} as 
\begin{equation}
    \hat v = v - \eta  \nabla f(x) / 2, \; \tilde x = x + \eta \hat v.
\end{equation}
If the new position $\tilde x$ is not in $K$,
then we reflect the HMC trajectory on the boundary. To achieve this, we assume that locally the trajectory of HMC is the segment $(1-t)x + t \tilde x$ that intersects $\partial K$ at a facet with normal vector $a_i$. We reflect the velocity as  $\mathrm{refl}_v(\hat v) = - 2 (\hat v^\top a_i) a_i + \hat v$ and then we reflect the position as $\mathrm{refl}_x(\tilde x) = \eta \cdot \mathrm{refl}(\hat v) + x$. We apply the reflection operator sufficiently many times 
until we obtain a position inside $K$; let the corresponding state be $(\hat v', \tilde x')$. 
Finally, we update the velocity, 
\begin{equation}
    \tilde v' = \hat v' - \eta \nabla f(\tilde x') / 2,
\end{equation}
we get the final state $(\tilde x', \tilde v')$, and we apply the Metropolis filter to transition from $(x, v)$ to $(\tilde x', \tilde v')$. 

\textbf{Cost per step.} 
Our implementation of ReHMC performs ---after a preprocessing--- the first reflection of a step in $O(Md)$ operations and each one of the rest reflections in $O(M)$ operations. Moreover, the integrator can be run for $w$ steps before proposing the new position. The parameter $w$ is called the \emph{walk length} and thus, the amortized per-step complexity becomes $O(M(d + \ell)w)$. The preprocessing step involves the computations of all inner products $a_i^\top a_j$ between the normal vectors of the facets, that takes $M^2 d$ operations. Now let $\hat{v}_j$ the velocity and $\tilde{x}_j,\ 1\leq j \le \ell$ the position before each reflection during a single step with $\hat{v}_1=\hat{v}$ and $\tilde{x}_1 = x$. During the computations of the first reflection we store all the values of the inner products $a_i^\top \tilde{x}_1$ and $a_i^\top \hat{v}_1$. For $j>1$, to compute the intersection time with $\partial K$, we pick the smallest positive root from the following linear equations,
\begin{equation*}
\begin{split}
    & a_i^\top((1-t_j)\tilde{x}_j + t_j(\tilde{x}_j + \eta\hat{v}_j)) = b_i,\ \text{where }t_j\in [0,1],\\
    & \tilde{x}_j = \tilde{x}_{j-1} + \eta\hat{v}_{j-1}\text{ and } \hat{v}_j = \hat{v}_{j-1}-2(\hat{v}_{j-1}^\top a') a',
\end{split}
\end{equation*}
and $a'$ is the normal vector of the facet that the trajectory hits at reflection $j-1$ and
$t_{j-1}$ the solution of the reflection $j-1$.  
We solve all the $M$ equations in $O(1)$ operations based on our bookkeeping from the  previous reflection and the preprocessing. When all the equations are infeasible we set $\tilde{x}' = \tilde{x}_j + \eta\hat{v}_j$.

\subsection{Mixing Time Analysis}

The crux of the matter of an MCMC algorithm is a bound on its  \emph{mixing time}. 
Roughly speaking, to sample from a target distribution with density $\pi$, we apply  (successively) a transition operator $\mathcal{T}$ to an initial density $\pi_0$ to obtain the distribution $\pi_k = \mathcal T^k \pi_0$, which approaches $\pi$ as $k \to \infty$. To measure the total variation distance between $\pi_k$ and $\pi$ within some accuracy $\varepsilon$ we consider the mixing time $\tau_{\mathrm{mix}}(\varepsilon ; \pi_0) = \inf \left \{ k \ge 0 \mid \| \mathcal T^k \pi_0 - \pi \|_{TV} \le \varepsilon \right \}$, that measures the number of iterations such that a sample $x_k \sim \pi_k$\footnote{If the chain is periodic $\pi_k \not \to \pi$, and the average distribution $\nu_k$ can be used~\cite{kannan2006blocking}.} for $k \ge \tau_{\mathrm{mix}}(\varepsilon; \pi_0)$ is within $\varepsilon$-TVD from a sample from $\pi$.  


\textbf{Assumptions.} To analyze the algorithm's performance  we make the following assumption: The step size $\eta$ is such that the sampler does at most $\ell$ reflections at each iteration. 

\textbf{Blocking Conductance.} The blocking conductance framework was introduced in \cite{kannan2006blocking} to address the pessimistic behaviour of \cite{Jerrum88} for the mixing time of a walk and to eliminate the ``start penalty'' that \cite{Jerrum88} proposes. This framework uses the idea of \emph{mixweight functions}\footnote{For a complete explanation of a mixweight function please refer to \cite{kannan2006blocking}.} to bound the total variation distance between $\nu_k$ and $\pi$. The work of \cite{lee2020logsmooth} determines a mixweight function that results   

\begin{theorem}[Blocking Conductance] \label{theorem:conductance} Let $\Pi_0$ (with density $\pi_0$) be a $\beta$-warm start for $\Pi$ with density $\pi$ both with common convex body support $K$. Suppose that for some $c_0$ and for all $c_0 \le t \le 1/2$ we have a bound of the form $\Pi(S) / \mathcal Q^2(S) \le \phi(t)$ for all $S \subset K$ with $\Pi(S) = t$, for a decreasing function $\phi$ on the range $[c_0, 1/4]$ with $\phi(t) \le M$ for $x \in [1/4, 1/2]$. Then
\begin{equation}
    \| \nu_k - \pi \|_{TV} \le \beta c_0 + \frac {32} k \left ( \int_{c_0}^{1/4} \phi(x) d x + \frac M 4 \right )
\end{equation}
\end{theorem}

Moreover, there is a bound on $\phi(t)$ defined for sets $\Omega \subset K$ that have high probability mass, that is
\begin{lemma}[Lemma 4.3 of \cite{lee2020logsmooth}] Let $\Pi$ be an $m$-strongly log-concave distribution with support the convex body $K$ and let $\Omega \subset K$ such that $\Pi(\Omega) = 1 - s$, for all $x, y \in \Omega$ we have that $\| \mathcal T_x - \mathcal T_y \|_{TV} \le 1 - a$, we have that $\eta \sqrt m < 1$ and $s \le \eta \sqrt m t / 16$. Then for all $t \in [0, 1/2]$ and $S \subseteq K$ with $\Pi(S) = t$ we have that 
\begin{equation}
    \phi(t) = \frac {2^{16}} {a^2 \eta^2 m t \log (1 / t)}.
\end{equation}
\end{lemma} 

This mixweight bound is based on a log-isoperimetric inequality from \cite{dwivedi2019log} and its application on the conductance bound yields a doubly logarithmic dependency on $\beta$. In addition, $\| \nabla f \| $ is concentrated around its mean, which provides a high-probability set. The bounds in \cite{lee2020logsmooth}, in general, are for an un-truncated log-concave density. However, 
they can be extended to truncated densities. This is so because a truncation of a density to a set $K$ is the product of an untruncated log-concave density with the indicator function. The latter is a log-concave function and therefore the product is a log-concave density as well.  

To derive our result we need to fill in the hypotheses of the corresponding Theorems. More specifically, we need to (i) determine a warm start, (ii) devise a step-size such that for ``close enough'' points we can state that the total variation distance between these points is bounded to be strictly less than 1. 

\textbf{Warm Starts.} We start by determining a warm start for our chain. Inspired by the untruncated case \cite{chen2019fast, dwivedi2019log, lee2020logsmooth}, our initial density is $\pi_0 = \mathcal N_K \left ( x^*, \tfrac{1}{L} I_d \right )$ where $\mathcal N_K$ denotes the \emph{truncated Gaussian density} on $K$, with mean the minimizer $x^*$\footnote{The minimizer can be determined using constrained optimization methods such as projected gradient descent or Frank-Wolfe methods \cite{frank1956algorithm}.} and variance $\tfrac{1}{L} I_d$. Then, the following holds for the warmness of $\pi_0$.
\begin{lemma}[App. \ref{sec:warm}] \label{lemma:warm}
    Let $\Pi$ be a log-concave distribution with density $\pi(x) \propto e^{-f(x)}$ defined on a convex body $K$, where $f$ is $L$-smooth, $m$-strongly convex, with condition number $\kappa = L / m$ and minimizer $x^* \in K^o$. If $\gamma$ is the sandwiching ratio of $K$, then the density $\pi_{0} = \mathcal N_K \left ( x^*, \tfrac 1 L I_d \right )$ is $O (\gamma^d \kappa^d)$-warm with respect to $\pi$. 
    
\end{lemma}

We can also initialize the sampler from a ``proxy start'' \cite{chen2019fast} when there is access to a point $z$ such that $\| x - z \| \le \delta$, for some $\delta > 0$, and an overestimation of the Lipschitz constant $\Lambda = (1 + \varepsilon_L) L$, for $\varepsilon_L > 0$. The proxy start is an $O (\gamma_z^{d} ((1 + \varepsilon_L) \kappa)^{d} \exp((\Lambda + m /2)\delta^2)$-warm, where $\gamma_z$ is the sandwiching ratio w.r.t.\ $z$ (App. \ref{sec:proxy}).

\textbf{Total Variation Bounds.} 
To establish a conductance bound, one has to bound the total variation distance between $\mathcal T_x$ and $\mathcal T_y$, for $x$ and $y$ being ``close starting points''. The analysis requires establishing a bound following the logic of applying the triangle inequality as  $\| \mathcal T_x - \mathcal T_y \|_{TV} \le \| \mathcal P_x - \mathcal P_y \|_{TV} + \| \mathcal T_x - \mathcal P_x \|_{TV} + \| \mathcal T_y - \mathcal P_y \|_{TV}$. Then bounding $\| \mathcal P_x - \mathcal P_y \|_{TV}$ requires bounding the total variation distance between mixtures of Gaussians by their KL Divergence. Bounding $\| \mathcal T_x - \mathcal P_x \|_{TV}$, and similarly $\| \mathcal T_y - \mathcal P_y \|_{TV} $, requires bounding the change in the energy $\mathcal H$. The technical difficulties we are facing to obtain these bounds are due to the reflections. From a bird's eye view, when a reflection occurs, the newly proposed state is not normally distributed, but is defined by a mixture of Gaussians. More specifically, given a sequence of normals at which the particle reflects at, the conditional distribution given the sequence of normals is a normal variable itself, since the composition of a sequence of linear operations and projections, on which the reflection operators are based, preserves Gaussianity. For this we need to bound the KL Divergence to subsequently bound $\| \mathcal P_x - \mathcal P_y \|_{TV}$. 

\begin{lemma} \label{lemma:kl_mixtures}
    Let $a, b: K \to [0, 1]$ be two probability density functions and let $p(x), q(x)$ be the densities $p(x) = \int_K a(y) \nrv {x | \mu_p(y)} {\eta^2 I_d} d y$, and $q(x) = \int_K b(y) \nrv {x | \mu_q(y)} {\eta^2 I_d} d y$
   where $\sup_{x, y \in K} \| \mu_p(x) - \mu_q(y) \| \le M$ for some $M \ge 0$. Then the KL divergence between $p$ and $q$ obeys the inequality $d_{KL} (p, q) \le \tfrac {M^2} {2 \eta^2} + d \log \left ( \eta^3 (2 \pi)^{3/2} e^{1/2} \right )$. If $\eta \le (2 \pi)^{-1/2} e^{-1/6}$, then $d_{KL}(p, q) \le \tfrac {M^2} {2 \eta^2}$. Moreover, if the conditional distributions have diagonal covariances with eigenvalues in the range $[\eta^2, (\ell + 1)^2 \eta^2]$ and $\eta \le \frac {e^{-(\ell + 1)^2 / 6}} {(2 \pi e)^{1/2} (\ell + 1)}$ we have that $d_{KL}(p, q) \le \frac {M^2}  {2 \eta^2}$. 

\end{lemma}

We use the bound on KL divergence to bound $\| \mathcal P_x - \mathcal P_y \|_{TV}$ by utilizing Pinsker's inequality.

\begin{lemma} \label{lemma:px_py}
        Let $\tilde x', \tilde y'$ be two points which are proposals concluded from points $x, y \in \Omega$ with $\| x - y \| \le \eta$ and $\Omega \subseteq K$ by executing one step of ReHMC (the \texttt{LEAPFROG} and \texttt{REFLECT} functions of Algorithm \pref{alg:hmc}) allowing at most $\ell \in \mathbb N^*$ reflections. If $\mathcal P_{x}, \mathcal P_y$ are the corresponding proposal distributions and $\eta \le \frac {1} {\sqrt {Lc} d (\ell + 1) \log (\kappa / \varepsilon)}  \le \frac {e^{-(\ell + 1)^2 / 6}} {(2 \pi e)^{1/2} (\ell + 1)}
        $ then $\sup_{\| x - y \| \le {\eta} } \| \mathcal P_x - \mathcal P_y \|_{TV} \le \frac 1 2 \left ( 1 + \tfrac {1} {2 c} \right )$.
    
\end{lemma}

\begin{table*}[t]
\centering

\scriptsize
\input{results}

\caption{Experimental Results for sampling from $\pi(x) \propto e^{ \| x - x_c \|^2 / 2}$ using ReHMC, Hit-and-Run from HOPS \cite{jadebeck2020hops} (H\&R-HOPS) and Coordinate-Hit-and-Run from HOPS (CH\&R-HOPS). The HOPS library uses H\&R (and CH\&R) together with an initial rounding procedure in order to sample from a convex polytope. The quantity $\psi_{\mathrm{me}}$ denotes the ratio of the maximum over the minimum axis lengths of the maximum-volume inscribed ellipsoid of $K$. A value of $\psi_{\mathrm{me}} \approx 1$  indicates that the body is in John's position \cite{john2014extremum}. The $\dagger$ symbol denotes a failed experiment where the sampler repeatedly escaped $K$, and the $\star$ symbol denotes failure to mix according to the $\text{PSRF} \le 1.2$ criterion. In each case, the most efficient sampler is highlighted (lower is better).}
\label{tab:results}
\end{table*}

Similarly, we derive bounds for $\| \mathcal T_x - \mathcal P_x \|_{TV}$ and $\| \mathcal P_y - \mathcal T_y \|_{TV}$. These bounds rely on bounding the change in the Hamiltonian $\mathcal H$ between the initial and the final positions.

Let $\Omega$ be the following set
\begin{equation} \label{eq:omega}
    \Omega \sdef \left \{ \omega \in K | \| \nabla f(\omega) \| \le \sqrt {L d} C \log (\kappa / \varepsilon) \right \}.
\end{equation}
We show that for every point $x \in \Omega$ the following Lemma holds
\begin{lemma}[App. \ref{sec:px_tx}] \label{lemma:px_tx}
    Let $x \in \Omega$ where $\Omega$ is as in \eqref{eq:omega}, and
    \begin{equation}
        \eta \le \frac {1} {\sqrt {Lc} d (\ell + 1) \log (\kappa / \varepsilon)}  \le \frac {e^{-(\ell + 1)^2 / 6}} {(2 \pi e)^{1/2} (\ell + 1)}.
    \end{equation}    
    Moreover let $(\tilde x', \tilde v')$ be the new proposal, where the sampler does $k \le \ell$ reflections, and let $\mathcal P_x$ and $\mathcal T_x$ be the corresponding proposal and transition distributions, respectively. Then
    $\| \mathcal P_x - \mathcal T_x \|_{TV} \le 1 - \frac 9 {10} \exp(-O(\tau))$, where $\tau = C / \sqrt c < 1$. 
\end{lemma}

\textbf{Main Result.}
By combining all the previous results, we conclude that if we fix $C = 1$ and let $c$ be large enough, then we can bound, strictly the total variation distance (from above) by 1, i.e., there exists some $a \in (0, 1)$ such that $\| \mathcal T_x - \mathcal T_y \|_{TV} \le 1-a$, where $x, y \in \Omega$ with $\| x - y \| \le \eta$.  

The previous remark and Theorem~\ref{theorem:conductance} lead to:
\begin{theorem}[Main Result, App. \ref{sec:main_result}] \label{thm:main_result}
    The ReHMC algorithm with a step size $\eta \le \frac {1} {\sqrt {Lc} d (\ell + 1) \log (\kappa / \varepsilon)} \le \frac {e^{-(\ell + 1)^2 / 6}} {(2 \pi e)^{1/2} (\ell + 1)}$ mixes in $O(\kappa d^2 \ell^2 \log^2(\kappa / \varepsilon) \log (d \log(\kappa /  \varepsilon) + d \log (\gamma / \varepsilon)) \log (1 / \varepsilon))$ steps, given a starting point $x_0 \sim \mathcal N_K(x^*, \tfrac 1 L I_d )$. 
\end{theorem}

\textbf{Implications for Uniform Sampling.} A particularly interesting subproblem is sampling from the uniform density $\pi(x) \propto \mathbf 1 \{ x \in K \}$ which can be modeled as the limit when the variance $\sigma$ of a Gaussian with $f(x) = \tfrac {\| x \|^2} {2 \sigma^2}$ with $0 \in K$ tends to $\infty$. In this case we have that $\kappa = 1$ as well as $\nabla f(x) \to 0$ for all $x \in K$ yielding a mixing time of $O(d^2 \ell^2 \log(d \log(\gamma / \varepsilon)) \log^3 (1 / \varepsilon)$ for a billiard-based walk similar to the one posed in~\cite{gryazina2014random}\footnote{The difference in this algorithm with~\cite{gryazina2014random} is that the whole segment is reflected until it gets in, rather having to traverse a certain ``trajectory length''.}. 

\section{Implementation and Experiments} \label{sec:experiments}

\textbf{Implementation.} 
We provide an open-source scalable C++ implementation of the ReHMC algorithm for general densities with access to the negative log-probability $f$ and its gradient $\nabla f$~\cite{anonymous_2021_4459362}.
It has been tested to work with multiple OS. 
Our software employs \eigen~\cite{eigenweb} for linear algebra and Intel's MKL library \cite{mklIntel} for high-performance linear algebra operations\footnote{Eigen's interface provides plug-and-play functionality with MKL.}. Our implementation supports convex polytope as domains, given as an intersection of half-spaces. We the optimized version of the \emph{Cyrus-Beck} algorithm~\cite{cyrus1978generalized}, described in Section \ref{sec:algorithm}, to calculate the intersection of the leapfrog trajectory (typically a line) with the boundary of the polytope. We perfomed the experiments on a machine with 16GB of RAM and an Intel i7 CPU at 2.6GHz.

\begin{figure}[t]
    \centering
    \includegraphics[width=0.45\textwidth]{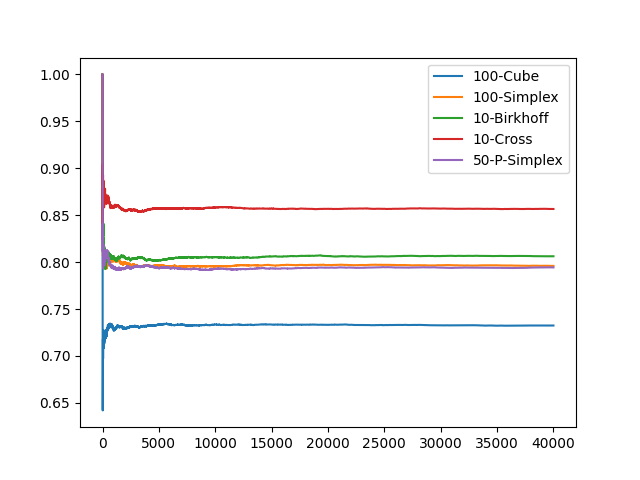}
    \caption{Adaptive step size rule behaviour for sampling from $\pi(x) \propto e^{-2 \| x - x_c \|^2 / R_c^2}$ where $x_c$ is the Chebyshev center and $R_c$ is the Chebyshev radius. We set $\eta_0 = R_c / 10$ and report the normalized step size sequence $\eta_t / \eta_0$. All sequences converge to certain values. A burn-in period of 40,000 iterations is shown, and convergence is observed at $\le 10,000$ iterations. 
    }
    \label{fig:adaptive_step_size}
\end{figure}

\textbf{Practical Parameterization of ReHMC.} Before sampling we perform a burn-in phase. Then, we pick the last point as a (warm) starting point for sampling. We exploit the steps we perform in the burn-in phase to learn an empirical value for the leapfrog step size. We compute a sequence $\eta_t$ of step sizes that converge to a value $\bar \eta$ in the long-run, which is the value of the step size we use for sampling. 
In particular, we use the following online rule:
We start from some initial value $\eta_0$ which we iteratively divide with the sample average number of reflections we have seen so far. More formally, let $\ell_1, \ell_2, \dots, \ell_t$ be the reflections observed until time $t$. Then, the step-size of time $t$ is $\eta_{t + 1} = \tfrac {\eta_t} {\left (\frac 1 t \sum_{j = 1}^t \ell_j \right )}$. We freeze the step-size after burn-in. Figure \ref{fig:adaptive_step_size} contains empirical evidence about the behaviour of $\{ \eta_t \}_{t \in \mathbb N}$.

\textbf{MCMC Diagnostics -- Evaluation.} To estimate the  practical efficiency of our method we measure the time needed to produce one independent sample after a total of $N$ draws, which we define as 
\begin{equation*}
    t_{\text{is}} = \frac {\text{Time to perform $N$ draws (us) }} {N_{\mathrm{ess}}} .
\end{equation*}
The Effective Sample Size $N_{\mathrm{ess}}$ (ESS) measures the amount by which autocorellation within chains increases uncertainty. Ideally, given independent, and hence uncorrelated, samples, the Central Limit Theorem outlines that the estimation error of the sample mean of the observations is $O(1 / \sqrt N)$. If there is correlation, then the estimation error is $O(1 / \sqrt {N_{\mathrm {ess}}})$. We use the definition and implementation of $N_{\mathrm{ess}}$ provided in \cite{geyer2011introduction} and compute the minimum (bottleneck) ESS among all dimensions. 

%
The logic behind reporting $t_{\mathrm{is}}$ is that the metric balances fast performance (i.e., the time needed to produce the next sample in the chain, which may be highly correlated with the previous one) and the ``bottleneck quality'' of sampling (i.e., a sampling algorithm may be slower but able to produce samples with lower correlation and hence higher $N_{\mathrm{ess}}$).

Moreover, we measure the Potential Scale Reduction Factor (PSRF) diagnostic that measures whether a chain has mixed by comparing the variance between and the variance within the chain components. We measure the PSRF of a chain by splitting it in half.

\textbf{Data.} We test our software on the following categories of convex polytopes:
\noindent 
\emph{1. Standard polytopes.} Such as, cubes, simplexes, cross-polytopes and products of simplexes (P-Simplex). \emph{2. Application polytopes.} Polytopes derived from BiGG models~\cite{bigg},  representing metabolic networks of biological systems.  
Their feasible regions correspond to the flux space of each  network that makes (flux) sampling a powerful tool to study metabolism~\cite{Herrmann19}.
    Furthermore, of special interest is the Birkhoff polytope, i.e.\ the convex hull of the set of permutation matrices, which has been widely used in the machine learning, computer vision and convex optimization communities for various permutation problems~\cite{Fogel13, Lim14}.

\textbf{Comparison Experiments.} We compare our approach with the standard sampling algorithm for sampling from a truncated log-concave density in practice which HOPS-H\&R and HOPS-CH\&R from the HOPS package, for sampling from a Gaussian density centered at the Chebyshev center with unit covariance. 
We run experiments with walk length $w \in [1, d]$ with an increment of $\lfloor d / 10 \rfloor$ for all samplers, while for ReHMC we set an initial step size equal to $\eta_0 = R_c / 10$ where $R_c$ is the Chebyshev radius.

For all methods we count $t_{\mathrm {is}}$ over a range of parameters regarding $w$, and report the minimum $t_{\mathrm{is}}$ observed given that the corresponding chain has a PSRF~$\le 1.2$. For ReHMC, we also report the average number of reflections for the selected experiment and the step size used \emph{after} burn-in. The results have been counted over a total of 80,000 draws per experiment with a burn-in of 20,000 draws and are presented in Table \ref{tab:results}. We apply a random rotation to 100-Cube, 100-Simplex, and 50-P-Simplex.


\textbf{Scaling Experiments.} To measure the scaling abilities of ReHMC we run sample from the following polytopes in the order of $\sim \! 10^3$ dimensions: 
(i) A 1000-Cube. 
(ii) a 1000-Simplex, 
(iii) the Birkhoff polytope with 33 elements ($d = 1024$), 
(iv) the computational biology polytope Recon1 ($d = 931, M = 4934$). 
We sample from a unit covariance Gaussian centered at the Chebyshev center for $N = 80,000$ draws with a burn in of 20,000 draws from a warm start. We report the results in Table~\ref{tab:scaling}.

\begin{table}[h]
    \centering
    \begin{tabular}{lrrrr}
    \toprule
        Polytope &  $w$ & $N_{\mathrm{ess}}$ & $t_{\mathrm{is}}$ (us) & PSRF \\
    \midrule 
        1000-Cube & 300 & 30903 & $3.9 \cdot 10^5$ & 1.001 \\
        1000-Simplex & 300 & 3257 & $1.18 \cdot 10^6$ & 1.010 \\
        33-Birkhoff & 300 & 25470 & $2.5 \cdot 10^5$ & 1.004 \\
        Recon1 & 187 & 409 & $1.2 \cdot 10^8$ & 1.021 \\
    \bottomrule
    \end{tabular}
    \caption{Scaling of ReHMC experiments.}
    \label{tab:scaling}
\end{table}

\section{Discussion}

\textbf{Competitors.} ReHMC is able to scale up to $1.03-1359\times$ faster than H\&R-HOPS. More specifically, while the two algorithms have close performances on the 100-Cube (ours surpassed H\&R by $1.03\times$), the H\&R algorithm was outperformed on the 100-Simplex ($45 \times$ faster), 100-S-Cube ($1.2\times$ faster),  10-Birkhoff ($8.8 \times$ faster), 10-Cross ($19.4\times$ faster), 50-P-Simplex ($31.9\times$ faster), e-coli ($1359.9\times$ faster), iAB-RBC-283, and iAT-PLT-636 ($\sim 22\times$ faster). The reason for the poor performance of H\&R on the simplex is its worst possible isotropic constant\footnote{The isotropic constant of a convex body $K$ is defined as $\det (\var {\mathcal U(K)} {x}) / \mathrm{vol}^2(K)$ where $\mathcal U(K)$ is the uniform distirbution.}  over all \emph{simplicial polytopes}~\cite{rademacher2016simplicial}. HOPS was also unable to sample from iAB-RBC-283 where it was not able to \emph{round} the polytope, due to its geometry.  Regarding the CH\&R-HOPS algorithm, it outperforms ReHMC only on the 100-Cube ($3.9 \times$),\footnote{The good performance of CH\&R is attributed to $\psi_{\mathrm{me}}$ being close to 1, the body being isotropic and the starting point being the center of the cube.} whereas it performs up to $386.1 \times$ slower on the rest of the benchmarks.  More specifically, we outperform CH\&R-HOPS on 100-Simplex ($8.9\times$ faster), 100-S-Cube ($4.07 \times$ faster), 10-Birkhoff ($26.7\times$ faster), 10-Cross ($5.23\times$ faster), 50-P-Simplex ($386.1 \times$ faster), iAB-RBC-283, and iAT-PLT-636 ($\sim 10 \times$ faster). 

Interestingly, ReHMC computes in all of our experiments a higher quality sampler, in terms of PSRF, than both H\&R and CH\&R. This could result in a larger performance gain for ReHMC if we restrict all samplers to stop after a certain PSRF value is attained.

\textbf{Scaling.} Moreover, ReHMC was able to scale up to thousands of dimensions and sample with very low PSRF, whereas contemporary implementations of \emph{truncated HMC}~\cite{afshar2015reflection} experimented with $\le 50$ dimensions. More specifically, we were able to efficiently sample (in the order of a few \emph{hours}) from a  1000-dimensional cube and simplex, a large Birkhoff polytope, and Recon1 metabolic model polytope.     

\textbf{Billiard Behaviour.} The average number of reflections $\bar \ell = \sum_{t = 1}^{Nw} \ell_t / (Nw)$ for every experiment was observed to have \emph{ergodic behaviour} across the range of $w$. 
Of course, as expected, the average number of reflections per-step varies from polytope to polytope with iAB-RBC-283 and 100-Simplex having a relatively high average number of reflections per step ($23.9943$ reflections, and 6.13 reflections per step on average respectively). 
Moreover, the number of reflections is expected to have dependence on $d$. We make the following \emph{conjectures} about the number of reflections: With appropriately chosen step size $\eta$, ReHMC does $\bar \ell = O(1)$ reflections when sampling from a standard Gaussian on $K = [-1, 1]$. When the domain is a cube $K = [-1, 1]^d$, then ReHMC does (by virtue of the union bound) $\bar \ell \propto d$ reflections and when the body is an $L_2$ ball, the average number of reflections drops to $\bar \ell \propto \sqrt d $. When the density has a condition number $\kappa$ and the polytope has sandwiching ratio $\gamma$ then $\bar \ell$ is minimized when $\gamma = \Theta (\kappa)$.    

\section{Conclusions, Impact \& Future Work}

We introduce an algorithm for sampling from a truncated log-concave density $\pi(x) \propto e^{-f(x)}$ using ReHMC. We analyze ReHMC and prove novel bounds about its mixing time when an off-line step-size is used. We also provide an online rule to estimate the step-size during the burn-in phase and we are able to  sample in practice from a variety of polytopes up to $\sim 10^3 \times$ faster than H\&R subject to empirical mixing criteria. While we believe that our general bound is not tight\footnote{It is tight with respect to the linear dependency on $\kappa$ with similarly to Thm. 4 of~\cite{chen2019optimal}.}, the question of whether it is tight as a function of $d$ and $\ell$ remains an interesting  direction for future research. Also, we leave as future work to employ different integrators to approximate the Hamiltonian dynamics and provide new mixing time guarantees. It is of special interest to compare them  through  extensive experiments. Impact-wise, the problem we investigate is mainly of theoretical nature with standard applications and poses no ethical considerations. 

\newpage

\bibliographystyle{icml2021}
\bibliography{icml_2021_paper.bib}

\newpage

\onecolumn {
\appendix
\begin{center}
    \Large
    \textbf{Supplementary Material}
\end{center}

\textbf{Note about notation.} In the Appendix, we make use of the notation $\exp (x)$ to denote $e^x$. The inequality $a \lesssim b$ denotes inequality up to a (universal) constant factor. 

\section{Markov-Chain-Monte-Carlo}

\subsection{Markov-Chain-Monte-Carlo  Algorithms} \label{sec:mcmc}

A very large family of algorithms in the sampling regime are the \emph{Markov-Chain-Monte-Carlo} (MCMC) (or \emph{Metropolis-Hastings}) algorithms introduced in the seminal works of \cite{metropolis1953equation, hastings1970monte}. The logic of an MCMC algorithm is the following: We start with an \emph{initial density} $\pi_0$ and we simulate the two following steps: First of all, we have a proposal step. The proposal step proposes a state $\tilde x$ given that the sampler is already in state $x$. The proposal step makes use of the proposal function $\mathcal P: K \times K \to [0, \infty)$, where $\mathcal P(x, \cdot)$ represents a density over $x \in K$. So, at the proposal step we sample a state $\tilde x \sim \mathcal (x, \cdot)$, and write $\tilde x \sim \mathcal P_x$ in shorthand. In the second step, known as the \emph{accept-reject} step, the algorithm accepts the proposal $\tilde x$ of the first step as the new state of the sampler with probability 

\begin{equation}
    \alpha(x, \tilde x) \sdef \min \left \{ 1,  \frac {\pi(\tilde x) \mathcal P(\tilde x, x)} {\pi(x) \mathcal P(x, \tilde x)}\right \}
\end{equation}

otherwise, with probability $1 - \alpha(x, \tilde x)$ the sampler rejects $\tilde x$ and remains in $x$. This process is also known as the Metropolis Filter \cite{neal2011mcmc} and is applied in order to ensure that $\pi$ is a stationary density for this Markov Chain.  Drawing the proposal $\tilde x$ and applying the accept-reject step can be combined to give the overall transition kernel $\mathcal T(x, \tilde x)$ defined as 

\begin{equation}
    \mathcal T(x, \tilde x) \sdef \mathcal P(x, \tilde x) \alpha(x, \tilde x) \qquad x \neq \tilde x
\end{equation}

where again $\mathcal T(x, \cdot)$ is a probability density function which we will denote in shorthand as $\mathcal T_x$ and thus $\tilde x \sim \mathcal T_x$. 

Sampling from a log-concave density can be performed in multiple ways, since the proposal distribution $\mathcal P_x$ can vary between the methods. Some methods include: (a) independence sampling where $\tilde x \sim \nrv 0 \Sigma$, (b) random-walk Metropolis (RWM) where $\tilde x \sim \nrv x {2 \eta I_d}$, (c) Metropolis-Adjusted Langevin Algorithm (MALA) where $\tilde x \sim \nrv {x - \eta \nabla f(x)} {2 \eta I_d}$, (d) Ball-walk (BW), (e) H\&R, (f) Coordinate-Hit-and-Run (CHR), (g) Underdamped Langevin Diffusion (ULD), and Hamiltonian (or Hybric) Monte Carlo (HMC). For a more detailed discussion of the various samplers we redirect the interested reader to \cite{chen2019fast, vempala2005geometric, chen2019fast, roberts2001inference, meyn2012markov, talay1990expansion, shen2019randomized} and the references therein.

\subsection{Hamiltonian Dynamics} \label{sec:hamiltonian_dynamics}

The Hamiltonian Dynamics \cite{betancourt2017conceptual, neal2011mcmc} is an interpretation for studying the evolution of physical systems. In this formulation, we have a particle of mass $m_p$ with velocity $v$ and position $x$. The particle moves in a conservative potential $\mathcal U(x)$ where it experiences a force $-\nabla \mathcal U(x)$ which is dependent from to its position and has a Kinetic Energy $\mathcal K(v) = \frac 1 2 m_p \| v \|^2$. The dynamics of the particle evolve according to Newton's Second Law, that is $m_p \dot v = - \nabla \mathcal U(x)$, or equivalently, in terms of the Hamiltonian $\mathcal H(x, v) = \mathcal K(v) + \mathcal U(x)$

\begin{align} \label{eq:hmc}
    \frac {d x} {d t} = +\frac {\partial \mathcal H} {\partial v} =  + v, \quad 
    \frac {d v} {d t} = -\frac {\partial \mathcal H} {\partial x} = - \frac 1 m_p \nabla \mathcal U (x)
\end{align}

The above system of equations preserves the Hamiltonian over time since

\begin{equation}
    \frac {d H} {d t} = \frac {\partial \mathcal H} {\partial x} \frac {d x} {d t} + \frac {\partial H} {\partial v} \frac {d v} {d t} = \frac {\partial \mathcal H} {\partial x} \frac {\partial H}{\partial v} - \frac {\partial H} {\partial v} \frac {\partial H} {\partial x} = 0
\end{equation}

Therefore the system moves on the level sets of the Hamiltonian function, that is $\mathcal C(E) = \{ (x, v) | \mathcal H(x, v) = E = \text{const.} \} $.  From now on, and for notational convenience we will assume that the particle has unit mass, that is $m_p = 1$. An alternative formulation of the Hamiltonian Dynamics defines a joint variable $z = (x, v)$ and evolves according to 

\begin{equation}
    \frac {d z} {d t} = J \nabla H(z), \quad J = \begin{pmatrix} O & I \\ -I & O \end{pmatrix}
\end{equation}

Or in terms of a mapping we can define $\mathbb T_s$ such that given a state $z(t)$, it produces the state $z(t + s)$ as 

\begin{equation}
    z(t + s) = \mathbb T_s z(t) = z(t) + \int_{t}^{t + s} J \nabla H(z(t)) d t
\end{equation}

with an inverse mapping $\mathbb T_{-s}$, which can be obtained by negating $v$, applying the forward mapping, and negating $v$ again. The Hamiltonian Dynamics are \emph{symplectic}, that is for the Jacobian mapping $B_s$ of $\mathbb T_s$ we have that $B_s^\top J^{-1} B_s = J^{-1}$. As a consequence, the operator is volume preserving as well, which means that for every region $R \subseteq \mathbb R^d$ we have $\mathrm {Vol}(R) = \mathrm {Vol} (\mathbb T_s R)$. This property can be proven by proving that the Jacobian of the mapping $\mathbb T_s$ for infinitesmall $s$ has absolute value 1, or the divergence of the vector field $F(z) = J \nabla H(z)$ is 0. 

\paragraph{Discretization of the Hamiltonian Dynamics.} Solving the Hamiltonian Dynamics ODE in a computer setting requires discretizing the underlying ode $\dot z = J \nabla H(z)$. For this reason, multiple methods have been proposed. The easiest one is, perhaps, the Euler method where

\begin{equation}
    v_{i + 1} = v_i - \eta \nabla f(x_i), \quad x_{i + 1} = x_i + \eta v_i
\end{equation}

and its improvement which uses the already computed value of $v_{i + 1}$

\begin{equation}
    v_{i + 1} = v_i - \eta \nabla f(x_i), \quad x_{i + 1} = x_i + \eta v_{i + 1}
\end{equation}

These methods, which are very simple and conceivable, are usually prone to numerical errors and may become unstable. Moreover, they have an $O(\eta^2)$ local error and an $O(\eta)$ global error. A better way to discretize the Hamiltonian Dynamics is through the \emph{leapfrog integrator}

\begin{equation}
    \hat v_{i + 1} = v_i - \frac \eta 2 \nabla f(x_i), \quad x_{i + 1} = x_i + \eta \hat v_{i + 1}, \quad v_{i + 1} = \hat v_{i + 1} - \frac \eta 2 \nabla f(x_{i + 1})
\end{equation}

which has an $O(\eta^3)$ local error and an $O(\eta^2)$ global error. An even smaller error, at the expense of computational power can be achieved with Runge-Kutta \cite{li2019stochastic, bou2017geometric}, Runge-Kutta-Nystr\"om , Verlet integrators \cite{monmarche2020high}, and Collocation Methods \cite{lee2018algorithmic}.

\subsection{Hamiltonian Monte Carlo} \label{sec:hmc}

Having gained intuition about the properties of the Hamiltonian Dynamics, we describe the Hamiltonian Monte Carlo Algorithm (HMC) \cite{duane1987hybrid, dang2019hamiltonian}. More specifically, HMC relies in simulating a particle $(x, v)$ with Kinetic Energy $\mathcal K(v) = \frac 1 2 \| v \|^2$ and Potential Energy $\mathcal U(x) = f(x)$ to draw samples from a target distribution $\Pi$. The state $z = (x, v)$ of the system evolves via the Hamiltonian Dynamics of \eqref{eq:hmc}. The sampler starts initially with a sample $x_0 \sim \Pi_0$ where $\Pi_0$ is the starting distribution (whose form determines how many iterations the algorithm does to mix to the desired distribution $\Pi$) and an initial velocity $v_0 \sim \nrv 0 {I_d}$ and runs an iteration, using a numerical integration method, to yield a proposal $(\tilde x_0, \tilde v_0)$. In the continuous setting, the Hamiltonian is preserved over time, which can be directly deduced using the chain rule and \eqref{eq:hmc}. When, however the ODE is solved with a computer, a discretization error is added and the Hamiltonian is not constant in general. For this reason, the sampler either sets  $x_1 $ equal to $ \tilde x_0$ with probability equal to  $\min \left \{ 1, \exp \left ( \mathcal H(x_0, v_0) - \mathcal H(\tilde x_0, \tilde v_0) \right ) \right \}$, or rejects the proposal with probability $1 - \min \left \{ 1, \exp \left ( \mathcal H(x_0, v_0) - \mathcal H(\tilde x_0, \tilde v_0) \right ) \right \}$, thus setting the sample $x_1$ again to $x_0$. The procedure repeats, generating sample $x_{i + 1}$ starting from the previous sample $x_i$ and a velocity $v_i \sim \nrv 0 {I_d}$. In the case of $K = \mathbb R^d$ (unconstrained sampling) we can use the second-order Leapfrog Integrator 
and return the proposal $(\tilde x, \tilde v)$ for some input $(x, v)$. This procedure leaves the joint distribution $\pi(x, v) \propto \exp(-\mathcal H(x, v))$ invariant. For every ``small'' set $A \subseteq \mathbb R^d$ and set $B$ reachable by $A$ through $\mathbb T_s$, we have that the Hamiltonian is constant (over an adequately small A) $\Pi(A) = \frac V Z \exp(-H_A)$, $\Pi(B) = \frac V Z \exp(-H_B)$ and 

\begin{equation}
    \frac V Z \exp(-H_A) \min \{ 1, \exp (-H_B + H_A \} = \frac V Z \exp(-H_B) \min \{ 1, \exp (-H_A + H_B) \}
\end{equation}

where $V = \mathrm {Vol}(A) = \mathrm {Vol}(B)$. For a more detailed introduction to the subject we redirect the interested reader to \cite{neal2011mcmc} and \cite{betancourt2017conceptual}. 


\section{Hamiltonian Monte Carlo for Truncated Sampling} \label{sec:truncated_hmc}

In Sections \pref{sec:hamiltonian_dynamics}, and \pref{sec:hmc} we have discussed the case where the potential $\mathcal U(x) = f(x)$ is a \emph{smooth function}, i.e. its gradient $\nabla f(x)$ does not explode at any point in the domain. In this section, we will focus on the setting where $\mathcal U(x)$ is non-smooth. More general, the form of $\mathcal U$ we assume is the following 

\begin{equation}
    \mathcal U(x) = 
    \begin{cases}
        f(x) & x \in K \\
        \infty & x \notin K
    \end{cases}
\end{equation}

where, again, $f$ is a an $L$-smooth and $m$-strongly convex function defined with domain a superset of $K$, and $K$ is a convex body. A particle under such a potential, encounters an \emph{infinite-potential barrier} and never has the energy to overcome it. The behaviour of this particle is therefore \emph{reflective} at the boundary. 



In the sampling context, the problem of sampling from this density is equivalent to sampling from

\begin{equation}
    \pi(x) \propto \exp (-\mathcal U(x)) \propto \begin{cases} \exp(-f(x)) & x \in K \\
    0 & x \notin K \end{cases}
\end{equation}

For this type of dynamics we can prove that 

\subsection{Volume preservation and time reversibility of the Continuous dynamics} \label{sec:volume_preservation_ideal}

\begin{theorem} \label{thm:volume_preservation_ideal}
    The ideal continuous Hamiltonian Dynamics preserve the volume of a region $A \subseteq K$ and are time-reversible.
\end{theorem}

\begin{proof}
    
    \textbf{Volume preservation.} We will prove the case for $d = 1$, since the multidimensional case is a direct generalization of our claim. Moreover, we assume that the domain has one boundary at $x = 1$. We choose some small $\delta > 0$. Firstly, applying the map $\mathbb T_\delta$ takes $(x(t), v(t))$ and produces $(x(t + \delta), v(t + \delta))$. This map is proven to be volume preserving in \cite{neal2011mcmc}. Now, if $x(t + \delta) > 1$ we need to reflect $x(t + \delta)$ and $v(t + \delta)$ respectively to fall inside $K$. We do this via the reflection operator $\mathbb U_\delta$ which is defined (in this case) to be
    
    \begin{equation}
        v'(t + \delta) = - v(t + \delta), \quad x'(t + \delta) = - \delta v(t + \delta) + x(t + \delta)  
    \end{equation}
    
    If $(x(t + \delta), v(t + \delta)) \le 1$ we let the operator $\mathbb U_\delta$ to equal the identity operator that sets $x'(t + \delta) = x(t + \delta), v'(t + \delta) = v(t + \delta)$.  

    It is straightforward to verify that the Jacobian of the transformation is $\begin{pmatrix}
        1 & -\delta \\
        0 & -1 \\
    \end{pmatrix}
    $ which has a determinant of -1 in the case of reflection and +1 in the case of no reflection. Finally we do a flip in $v'(t + \delta)$ and the determinant of the flip together with $\mathbb U_\delta \circ \mathbb T_\delta$ becomes +1. For a larger trajectory $s$, we split it to $N$ pieces and set $\delta = s / N$. Applying the dynamics at each segment with step $\delta = s / N$ yields $k_1, \dots, k_N \in \mathbb N$ such that the trajectory of the particle is represented at $\mathbb U_{s / N}^{k_N} \circ \mathbb T_{s / N} \circ \mathbb U_{s / N}^{k_{N-1}} \circ \mathbb T_{s / N} \circ \dots \circ \mathbb U_{s/N}^{k_1} \circ \mathbb T_{s / N}$. It can be shown similarly to \cite{neal2011mcmc} that the absolute value of the log determinant is $O(1 / N)$. Therefore the reflective dynamics are volume preserving since as $N \to \infty$ the absolute value of the determinant converges to 1.  
        
    \textbf{Time reversibility.} The dynamics are time-reversible since inverting the operators $\mathbb T_\delta$ and $\mathbb U_\delta$ to $\mathbb T_{- \delta}$ and $\mathbb U_{-\delta}$ respectively, and running the dynamics with initial state $(x(t + \delta), v(t + \delta))$, yield $(x(t), v(t))$.

\end{proof}

\begin{algorithm}[t]

\caption{Leapfrog Integrator.}
\label{alg:hmc}
\small
\begin{algorithmic}

\Procedure{\texttt{Walk}}{$f, \eta, x, v, w$}
    \For {$1 \le i \le w$} 
        \State $(x, v) \gets$\text{\texttt{LEAPFROG}} ($f, \eta, x, v$)
    \EndFor
\EndProcedure

\\

\Procedure {\texttt{Leapfrog}}{$f, \eta, x, v$}
    \State $\hat v \gets v - \frac {\eta} 2 \nabla f(x)$
    \State $\tilde x \gets x + \eta \hat v$
        \If{$\tilde x \in K$}
        \State $\tilde x' \gets \tilde x$
        \State $\hat v' \gets \hat v$
    \Else
        \State $(\tilde x', \hat v') \gets \text{\texttt{REFLECT}}(K, \tilde x, x, \hat v)$
    \EndIf
    \State $\tilde v' \gets \hat v' - \frac {\eta} 2 \nabla f(\tilde x')$
    \State \Return $(\tilde x', \tilde v')$
\EndProcedure

\end{algorithmic}

\end{algorithm}

\begin{algorithm}[t]

\caption{Reflection operation.}
\label{alg:hmc}
\small
\begin{algorithmic}
\Procedure {\texttt{Reflect}}{$K, \tilde x, x, \hat v$}
    \State $a = \tilde x - x$
    \State $u$ is the point of intersection of $x + t a$ and $\partial K$
    \State $n$ is the normal vector at $u \in \partial K$
    \State $\tilde x' \gets -2 (a^\top n) n + a + \tilde x$
    \State $\hat v' \gets - 2 (\tilde v^\top n) n + \hat v $
    \If {$\tilde x' \notin K$} 
    \State \texttt{REFLECT}($K, \tilde x', \tilde x, \hat v'$)
    \Else
    \State \Return $(\tilde x', \hat v')$
    \EndIf
\EndProcedure

\end{algorithmic}

\end{algorithm}

\begin{algorithm}[t]

\caption{Hamiltonian Monte Carlo with Boundary Reflections.}
\label{alg:truncated_hmc}
\small
\begin{algorithmic}

\Procedure{\texttt{HMC}}{$\eta$, $f$, $K$, $N_{samples}, w$}
\State $k \gets 0$
\State $x^* \gets \text{\texttt{MINIMIZE}}(f, K)$
\State Draw $x_0 \sim \mathcal N_K(x^*, L^{-1} I_d)$
\While{$k \le N_{samples}$}
    \State Draw $v_k \sim \nrv 0 {I_d}$ 
    \State $(\tilde x_k, \tilde v_k) \gets \text{\texttt{WALK}}(f, \eta, x_k', v_k', w)$
    \State Draw $u \sim \mathcal U[0, 1]$
    \If{$u \le \min \{1, \exp \left ( \mathcal H(x_k, v_k) - \mathcal H (\tilde x_k', \tilde v_k') \right ) \}$} 
    \State $x_{k + 1} \gets \tilde x_k'$
    \Else
    \State $x_{k + 1} \gets x_k$    
    \EndIf
    \State $k \gets k + 1$
\EndWhile
    \State \Return $\{ x_{k} \}_{1 \le k \le N_{samples}}$
\EndProcedure

\end{algorithmic}

\end{algorithm}

In the discretized Hamiltonian Dynamics with the leapfrog integrator we first perform the \emph{velocity half-update} and the \emph{position update} from an initial state $(x, v)$ as 

\begin{equation}
    \hat v = v - \frac \eta 2 \nabla f(x), \qquad \tilde x = x + \eta \hat v
\end{equation}

Note that the newly computed position $\tilde x$ may not lie in $K$. In case it does not lie inside $K$ we need perform a reflection as follows

\begin{equation}
    \hat v \mapsto - 2 (\hat v^\top n) n + \hat v, \qquad \tilde x \mapsto - 2 \eta (\hat v^\top n) n  + \eta \hat v + x
\end{equation}

where $n$ is the normal at the point of the intersection $\{ z | z = t x + (1 - t) \tilde x, t \in [0, 1] \} \cap K$, and the reflection can be applied multiple times until the position falls inside $K$ yielding the state $(\tilde x', \hat v')$. We then perform the final velocity step

\begin{equation}
    \tilde v' = \hat v' - \frac \eta 2 \nabla f(\tilde x')
\end{equation}

We can prove that these dynamics are volume preserving since we can break the transformation to 3 parts $(x, v) \mapsto (\tilde x, \hat v) \mapsto (\tilde x', \hat v') \mapsto (\tilde x', \tilde v')$ each of which is trivially volume preserving. We present the omitted proof Theorem \ref{thm:reflective_dynamics_volume_preserving}

\subsection{Volume preservation and time reversibility of the discretized dynamics} \label{sec:reflective_dynamics_volume_preserving}

\begin{theorem}
\label{thm:reflective_dynamics_volume_preserving}
    The discretized reflective Hamiltonian Dynamics are volume-preserving and time-reversible.
\end{theorem}

\begin{proof}
    \textbf{Volume preservation.} We will prove the theorem for $d = 1$ and assuming that the domain has a boundary at $x = 1$ (here the assumption that $K$ is bounded is not needed). The leapfrog dynamics map consists of the following two maps $\mathbb G_\eta, \mathbb H_\eta$ with 
    
    \begin{align}
        \mathbb G_\eta: & \qquad \hat v = v - \frac \eta 2 f'(x), \quad \hat x = x + \eta \hat v \\
        \mathbb H_\eta: & \qquad \tilde v = \hat v - \frac \eta 2 f'(\hat x), \quad \tilde x = \hat x
    \end{align}
    
    We also define the reflection operator $\mathbb U_\eta$ as 
    
    \begin{equation}
        \mathbb U_\eta: \qquad \hat v' = - \hat v, \quad \tilde x' = - \eta \hat v + \tilde x
    \end{equation}
    
    The volume preservation properties of $\mathbb G_\eta \circ \mathbb H_\eta$ have been proven analytically in \cite{neal2011mcmc}. The more general case that the reflective dynamics impose is the one of 
    
    \begin{equation}
        \mathbb G_\eta \circ \underbrace{\mathbb U_\eta \circ \dots \circ \mathbb U_\eta}_{\text{at most } \ell \text{ times}} \circ \mathbb H_\eta
    \end{equation}
    
    per iteration in the case of multiple boundary normals (trivially in the case of $x = 1$ being the only normal we have $\ell = 1$). The reflection map has been proven to be volume-preserving in Theorem \ref{thm:volume_preservation_ideal} therefore each iteration is volume preserving, so for each step the absolute value of the determinant of the transformation is 
    
    \begin{equation}
        \left | \det  \begin{pmatrix} 1 - \eta^2 /2 f''(x) & \eta \\ - \eta / 2 f''(x) & 1 \end{pmatrix}  \cdot \prod_{i = 1}^k \det \begin{pmatrix} 1 & -\eta \\ 0 & -1 \end{pmatrix} \cdot \det \begin{pmatrix} 1 & 0 \\ -\eta/2 f''(\tilde x') & 1 \end{pmatrix}\right  | = 1
    \end{equation}
    
    for some $k \in \{0, \dots, \ell \}$. Thus the dynamics are volume-preserving. 
    
    \textbf{Time reversibility.} The time reversibility of the dynamics can be proven by applying the operator sequence $
        \mathbb G_{-\eta} \circ \underbrace{\mathbb U_{-\eta} \circ \dots \circ \mathbb U_{-\eta}}_{\text{at most } \ell \text{ times}} \circ \mathbb H_{-\eta}
    $ to the proposed state $(\tilde x', \tilde v')$ to obtain the initial state $(x, v)$.
    
    \textbf{Multivariate case.} For $d \ge 1$ dimensions and $m = 1$ constraint, the determinant of the Jacobian of the operator $\mathbb U_\eta$ equals the determinant of $\begin{pmatrix} I_d & \eta (-2 nn^\top + I_d) \\ O_d & - 2 nn^\top + I_d \end{pmatrix}$, which has a value of 1 since $n$ is a unit normal vector. For $m \ge 1$ constraints the map $\mathbb U_\eta$ consists of submatrices of the previous form and hence has absolute determinant 1. With similar arguments we can calculate the Jacobian of $\mathbb G_\eta, \mathbb H_\eta$ and use Leibniz's rule to calculate the block determinants.  
    
\end{proof}

\subsection{Reflection Operations} \label{sec:reflection_operator}



When at most one reflection occurs, the trajectory between $x$ and $\tilde x$, namely $\{ z | z = t x + (1 - t) \tilde x, t \in [0,1] \}$ can lie inside $K$ or intersect with $K$ at a point $y \in \partial K$. Therefore, we can define a \emph{density function} $\alpha: K \to [0, 1]$ such that the proposal distribution $\mathcal P_x$ can be expressed as 

\begin{equation} \label{eq:gaussian_mixture}
    d \mathcal P_x(\tilde x') \propto \int_K \alpha(y) w(\tilde x' | y) d y 
\end{equation}

where $w(\cdot | y)$ is the probability density function of $\tilde x'$ conditioned on a reflection at $y$ with a normal vector $n(y)$ (or no reflection if the proposed position lies inside $K$). It is easy to observe that this distribution is again a Gaussian density, given by the following Lemma.

\begin{lemma} \label{lemma:reflection}
    Let $K$ be a convex body and let $x \in K$ be a known point and let $\tilde x \in \mathbb R^d$ be the proposed point of the Algorithm \ref{alg:hmc} by the \texttt{LEAPFROG} function. Then if $a = \tilde x - x$
    
    \begin{equation}
        a \sim \nrv {\frac {- \eta^2} 2 \nabla f(x)} {\eta^2 I_d}
    \end{equation}
    
    Moreover let $\tilde x'$ be a point such that it does at most one reflection, namely
    
    \begin{align}
        \tilde x' |_{\tilde x, n}  & = \begin{cases}
            \tilde x & \tilde x \in K \\
            -2 (a^\top n)n + a + \tilde x & \tilde x \notin K
        \end{cases}
    \end{align}
    
    Where $n$ is a known unit normal vector of $K$. Then 
    
    \begin{align}
        \tilde x' |_{\tilde x \not \in K, n} \sim \nrv {x - \eta^2 \nabla f(x) + \eta^2 (\nabla f^\top (x) n) n} {4 \eta^2 I_d}
    \end{align}
    
\end{lemma}

\begin{proof}
    To prove the first claim we directly refer to the algorithm since 
    
    \begin{equation}
        a = \tilde x - x = \eta v - \frac {\eta^2} {2} \nabla f(x) \implies a \sim \nrv { - \frac {\eta^2} 2 \nabla f(x)} {\eta^2 I_d}
    \end{equation}
    
    For the second part we observe that $\tilde x' |_{\tilde x \not \in K, n}$ is a result of operations on Gaussian variables hence it will be a Gaussian variable itself. We start by defining the variable
    
    \begin{equation}
        w \sdef a^\top n = \sum_{i = 1}^d a_i n_i \implies w \sim \nrv {\frac {-\eta^2} 2 \nabla f^\top (x) n} {\eta^2}
    \end{equation}

    We start by computing the expectation
    
    \begin{equation}
        \begin{split}
            \ev {v \sim \nrv 0 {I_d}} {\tilde x' | \tilde x \not \in K, n} 
            = \ev {} {-2w n + a + \tilde x}
            = x - {\eta^2}  \nabla f(x) + \eta^2 (\nabla f^\top (x) n) n
        \end{split}
    \end{equation}
    
    And then the variance 
   
     \begin{equation}
        \begin{split}
            \var {v \sim \nrv 0 {I_d}} {\tilde x' | \tilde x \not \in K, n} 
            = 4 \left ( \var {} {a} + \var {} {wn} - 2 \mathrm {Cov} (a, wn) \right ) 
            = 4 \eta^2 I_d
        \end{split}
    \end{equation}
   
    Where
    
    \begin{align}
        \var {} {a} = \eta^2 I_d, \quad 
        \var {} {2wn} = 4 \var {} {wn} = 4 \eta^2 n n^\top, \quad
        2 \mathrm {Cov} (a, wn) = 2 \mathrm {Cov} (a, wn) = 2 \eta^2 nn^\top
    \end{align}
    
    Since the $(i, j)$ element for $i, j \in [d]$ is 
    
    \begin{equation}
        \mathrm {Cov} (a_i, wn_j) = \mathrm {Cov} (a_i, w) n_j = n_j \sum_{k = 1}^d \mathrm {Cov} (a_i, a_k n_k) = n_j \sum_{k = 1}^d \mathrm {Cov} (a_i, a_k) n_k = n_j \sum_{k = 1}^d n_k \eta^2 \mathbf 1 \{ i = k \} = n_i n_j \eta^2
    \end{equation}
    
\end{proof}

In case of multiple reflections we continue in a similar manner. Namely, when $k \le \ell$ reflections are observed at unit normal vectors $n_1, \dots, n_k$, then the conditional distribution of $x$ given the reflections at $n_1, \dots, n_k$ is a normal random variable and have a weight that corresponds to the probability that the sequence of reflections $n_1, \dots, n_k$ is followed given the initial position of the sampler. Note that these events are \textbf{not} statistically independent. This is a very crucial point in our analysis, since the total variation bounds we extract for this algorithm involve bounding the total variation distance between mixtures of Gaussians which have the form of \eqref{eq:gaussian_mixture}.  

}

\section{Mixing Time Analysis} 

\subsection{Warm Starts} 

\subsubsection{Proof of Lemma \ref{lemma:warm}} \label{sec:warm}

\begin{proof}
    Recall that from $L$-smoothness and $m$-strong-convexity for $x \in K$ and $y = x^*$ we have that     
    
    \begin{equation}
        \frac m 2 \| x - x^* \|^2 \le f(x) - f(x^*) \le \frac L 2 \| x - x^* \|^2
    \end{equation}
    
    Equivalently, since $\exp(-t)$ is a decreasing function

    \begin{equation}
        0 \le \exp \left ( - \frac L 2 \| x - x^* \|^2 \right ) \lesssim \exp(-f(x)) \lesssim \exp \left ( - \frac m 2 \| x - x^* \|^2 \right )
    \end{equation}
    
    Integrating inside $K$ we have that

    \begin{equation} \label{eq:integrate}
        0 \le \int_K \exp  \left ( - \frac L 2 \| x - x^* \|^2 \right ) d x \lesssim \int_K \exp(-f(x)) d x \lesssim \int_K \exp \left ( - \frac m 2 \| x - x^* \|^2 \right ) d x
    \end{equation}

    We calculate the warmness function $\beta: K \to \mathbb (0, +\infty)$
    
    \begin{equation}
        \beta(x) = \frac {d \mathcal N_K (x | x^*, 1/L I_d)} {d \Pi (x)} = \frac {\exp \left ( - \frac L 2 \| x - x^* \|^2 \right )} {\exp (-f(x))} \cdot \frac {\int_K \exp(-f(z)) d z } {\int_K \exp \left ( - \frac L 2 \| x - x^* \|^2 \right ) d  x} = C \cdot \beta_1(x)
    \end{equation}
    
    From \eqref{eq:integrate} the above $\beta_1(x) \le 1$ for all $x \in K$. We now need to bound the constant 
    
    \begin{equation}
        C = \frac {\int_K \exp(-f(z)) d z } {\int_K \exp \left ( - \frac L 2 \| x - x^* \|^2 \right ) d x}
    \end{equation}
    
    Let $\mathfrak B_1 = \mathbb B(x^*, r), \mathfrak B_2 = \mathbb B(x^*, R)$ be the two balls with radii $0 < r < R$ respectively such that 
    
    \begin{equation}
        \mathfrak B_1 \subseteq K \subseteq \mathfrak B_2        
    \end{equation}

    and $\gamma = R / r \ge 1$ is the sandwiching ratio. It is direct from the properties of integrals on non-negative and non-zero everywhere functions that 
    
    \begin{equation}
        \frac {\int_{\mathfrak B_1} \exp(-f(z)) d z } {\int_{\mathfrak B_2} \exp \left ( - \frac L 2 \| x - x^* \|^2 \right ) d x} \le C \le  \frac {\int_{\mathfrak B_2} \exp(-f(z)) d z } {\int_{\mathfrak B_1} \exp \left ( - \frac L 2 \| x - x^* \|^2 \right ) d x}
    \end{equation}
    
    We are interested in the upper bound. Using strong-convexity again we have that 
    
    \begin{equation}
        C \le  \frac {\int_{\mathfrak B_2} \exp \left ( - \frac m 2 \| x - x^* \|^2 \right ) d x } {\int_{\mathfrak B_1} \exp \left ( - \frac L 2 \| x - x^* \|^2 \right ) d x}
    \end{equation}
    
    Doing a change of variables $u = \frac 1 {\sqrt m} (x - x^*)$ and $w = \frac {1} {\sqrt L} (x - x^*)$ where the volume elements become $d u = m^{-d/2} d x$ and $d w = L^{-d/2} d x$ since the absolute values of the Jacobians of the corresponding transformations are $m^{-d/2}$ and $L^{-d/2}$ respectively, and the transformed domains are $\mathbb B(0, R / \sqrt m)$ and $\mathbb B(0, r / \sqrt L)$ we arrive at the fact that 
    
    \begin{equation} 
        C = \kappa^{d/2} \frac {\int_{\mathbb B(0, R / \sqrt m)} \exp(-\| u \|^2 / 2) d u} {\int_{\mathbb B(0, r/\sqrt L)} \exp(-\| w \|^2 / 2) d w} \le \kappa^{d/2} \left ( \frac {\mathrm{erf}(R / \sqrt m)} {\mathrm{erf}(r / \sqrt L)} \right )^d < \kappa^{d / 2} \left ( \frac {1 -  \exp(-2R^2 / m)} {1 - \exp \left ( -r^2 / L \right )} \right )^{d / 2}
    \end{equation}
    
    by the well-known identity of the Gaussian integral in polar coordinates $\pi (1 - \exp(-a^2)) < \mathrm{erf}^2(a) < \pi ( 1 - \exp (- 2a^2) )$ 
    
    where $\mathrm{erf}(t) = \int_{-t}^t \exp(-z^2) dz$ (we ignore the constant $\sqrt \pi / 2$ in front of its official definition since we are interested in bounding a ratio of quantities involving the same constant). Using the fact that $R = \gamma r$, the relation $\exp(x) \ge 1 + x$ we have that $1 - \exp(- 2 R^2 / m) \le 2 R^2 / m = 2 \gamma^2 r^2 / m$. Moreover, using the Taylor series for $\exp(x) \approx 1 + x + O(x^2)$ for small $x$, we get that $1 - \exp(-r^2 / L) \approx r^2 / L + O(r^4 / L^2)$. The fraction in question can be therefore shown to behave asymptotically as
    
    \begin{equation}
        C \le \kappa^{d / 2} \left ( \frac {1 -  \exp(-2 \gamma^2 r^2 / m)} {1 - \exp \left ( -r^2 / L \right )} \right )^{d / 2} = O \left ( \kappa^{d / 2} \left ( \frac {2 \gamma^2 r^2 / m} {r^2 / L} \right )^{d / 2} \right ) = O((\kappa \gamma)^d)
    \end{equation}
    
    since in the worst case the smaller ball becomes very small (hence the Taylor expansion for the denominator). The Taylor approximation error is of the type of $1 / (1 + O(h)) \approx 1 - O(h)$ for small $h$.  
    
    
    
    Hence $\beta(x) = O((\kappa \gamma)^d)$. The lower bound can be achieved when the convex body is a ball centered at the minimizer, where the bound reduces to its previous form. 

\end{proof}

\subsubsection{Proxy Start} \label{sec:proxy}

There are cases however that we do not have access to the minimizer, i.e. the minimizer is placed on an ``unconvenient'' place like the boundary of $K$, or the actual smoothness parameter $L$ is not known and we have access to an estimate $\Lambda = (1 + \varepsilon_L) L$ for some $\varepsilon_L \ge 0$. In this case, we can use a ``proxy'' distribution

\begin{equation}
    \mathcal N_K \left (z, \frac {1} {2 \Lambda} I_d \right )
\end{equation}

in order to start our sampler from. We assume that for some $\delta > 0$ we have $\| x^* - z \| \le \delta$. We can then easily prove the following Lemma about the proxy start.

\begin{lemma}[Proxy start] \label{lemma:proxy} 
    The distribution 
    
    \begin{equation}
        \pi_{0}^{\mathrm{proxy}} = \mathcal N_K \left ( z, \frac {1} {2 \Lambda} I_d \right )
    \end{equation}
    
    is a $O (\gamma_z^{d} ((1 + \varepsilon_L) \kappa)^{d} \exp((\Lambda + m /2)\delta^2)$-warm distribution with respect to $\pi$, where $\gamma_z = \inf_{R > r > 0} \{ R/r | \mathbb B(z, r) \subseteq K \subseteq \mathbb B(z, R) \} $.
\end{lemma}

\begin{proof}
    By the triangle inequality (also appears in~\cite{chen2019fast}) we can deduce that
    
    \begin{equation} \label{eq:treq}
        \| x - z \|^2  \ge \frac 1 2 \| x - x^* \|^2 - \| x^* - z \|^2
    \end{equation}
    
    and 
    
    \begin{equation}
        \exp \left ( - \frac {\Lambda} 2 \| x - z \|^2 \right )  \le \exp \left ( \frac {\Lambda} 2 \delta^2 \right ) \exp \left ( - \frac L 4 \| x - x^* \|^2 \right ) 
    \end{equation}
    
    Also by exchange of $x^*$ and $z$ we can get 
    
    \begin{equation} \label{eq:treq2}
            \| x - x^* \|^2  \ge \frac 1 2 \| x - z \|^2 - \| x^* - z \|^2 
    \end{equation}
    
    and therefore
    
    \begin{equation}
            \exp \left ( - \frac m 2 \| x - x^* \|^2 \right ) \le \exp \left ( \frac m 2 \delta^2 \right ) \exp \left ( - \frac m 4 \| x - z \|^2  \right ) 
    \end{equation}
    
    We now follow the same procedure as in Lemma \ref{lemma:warm}
    
    \begin{equation}
        \begin{split}
            \frac {d \mathcal N_K \left (z, \frac 1 {2 \Lambda} I_d \right )} {d \Pi} & = \frac {\exp (- \Lambda \| x - z \|^2)} {\exp(-f(x))} \cdot \frac {\int_K \exp(-f(z)) d z} {\int_K \exp(-\Lambda \| w - z \|^2) d w} \\
            & \le \exp(\Lambda \delta^2) \cdot \frac {\exp (- L / 2 \| x - z \|^2)} {\exp(-f(x))} \cdot \frac {\int_K \exp(-f(z)) d z} {\int_K \exp(-\Lambda \| w - z \|^2) d w} \\
            & \le \exp((\Lambda + m /2) \delta^2) \cdot \frac {\int_K \exp(-m / 4 \| w - z \|) d w} {\int_K \exp(-\Lambda \| w - z \|^2) d w} \\
            & \le \exp((\Lambda + m /2) \delta^2) \cdot \frac {\int_{\mathbb B(z, R)} \exp(-m / 4 \| w - z \|) d w} {\int_{\mathbb B(z, r)} \exp(-\Lambda \| w - z \|^2) d w} \\
            & = O (\gamma_z^{d} ((1 + \varepsilon_L) \kappa)^{d} \exp((\Lambda + m /2)\delta^2)
        \end{split}
    \end{equation}
    
    Where the first inequality is due to \eqref{eq:treq}, the second inequality is due to \eqref{eq:treq2} and the last two inequalities follow the exact same proof technique that Lemma \ref{lemma:warm} does.
    
\end{proof}

The above lemma establishes the trade-off for moving the starting point and changing the Lipschitz constant with an over-estimate in terms of the sandwiching ratio around the proxy point $z$ and the new condition number which is an $(2 + 2 \varepsilon)$-factor apart from the original one. Moreover, the shifting from the minimizer position comes with an overhead of $\exp((\Lambda + m / 2) \delta^2)$. Samples from these truncated normal distributions can be obtained by using the algorithm of~\cite{cousins2015bypassing}.


\subsection{Total Variation Bounds}

\subsubsection{Proof of Lemma \ref{lemma:kl_mixtures}} \label{sec:kl_mixtures}

\begin{proof}
    \emph{Case 1: Equal covariance matrices} Our proof will be based on the utilization of the upper bound of \cite{durrieu2012lower} (Eq. 20) where we can replace sums with integrals and maintain correctness. Using Lemma \ref{lemma:entropy_gaussian} since all Gaussians have the same covariance matrices equal to $\eta^2 I_d$
    
    \begin{equation}
        \int_K a(y) \ev {x \sim \nrv {\mu_p(y)} {\eta^2 I_d}} {- \log \nrv {x | \mu_p(y)} {\eta^2 I_d}} d y = \frac {d} 2 \log (2 \pi e \eta^2) \\
    \end{equation}
        
    Furthermore from the fact that all mean distances are bounded above by $M$, coefficients are positive and sum up to 1, and that the KL divergence between any two condintional densities is at most $M^2 / 2 \eta^2$ we get that 
    
    \begin{equation}
        - \int_K a(y) \log \int_K b(z) \exp (-d_{KL} ( \nrv {\mu_p(y)} {\eta^2 I_d},  \nrv {\mu_p(z)} {\eta^2 I_d})) d z d y \le \frac {M^2} {2 \eta^2}
    \end{equation}
    
    The integral of the product of numerators  between any two conditional densities of $p$ centered at $\mu_p(x)$ and $\mu_p(y)$ respectively is bounded as
    
    \begin{equation}
    \begin{split}
        Z(x, y) & = \int_{\mathbb R^d} \exp \left ( - \frac {(u - \mu_p(x))^\top (u - \mu_p(x))} {2 \eta^2} \right ) \cdot \exp \left ( - \frac {(u - \mu_p(y))^\top (u - \mu_p(y))} {2 \eta^2} \right ) du \\
        & \le \int_{\mathbb R^d} \exp \left ( - \frac {\| u \|^2} {\eta^2} \right ) du \\
        & \le (2 \pi)^{d / 2} \det (\eta^2 / 2 I_d)^{1/2} \\
        & = (2 \pi)^{d / 2} (\eta^2 / 2)^{d / 2} \\
        & = ( \pi \eta^2)^{d / 2} \\
        & \le ( 2 \pi e \eta^2)^{d / 2}
    \end{split}
    \end{equation}
    
    where we have used the Cauchy–Bunyakovsky–Schwarz that states that for two square integrable real valued functions $\iota_1, \iota_2$ and with support $S$ we have that 
    
    \begin{equation*}
        \left | \int_S \iota_1(x) \cdot \iota_2(x) dx \right |^2 \le \int_S \iota_1^2(x) dx \cdot \int_S \iota_2^2(x) dx 
    \end{equation*}
    
    where we have set $\iota_1(u) = \exp \left ( - \frac {(u - \mu_p(x))^\top (u - \mu_p(x))} {2 \eta^2} \right )$, $\iota_2(u) = \exp \left ( - \frac {(u - \mu_p(y))^\top (u - \mu_p(y))} {2 \eta^2} \right )$ and have observed that due to this definition the squared functions correspond to the numerators of Gaussian densities centered at $\mu_p(x)$ and $\mu_p(x)$ with covariance $\eta^2 / 2 I_d$ and hence the two integrals are bounded by the normalization constant of $\nrv {0} {\eta^2 / 2 I_d}$ which is bounded above by $( 2 \pi e \eta^2)^{d / 2}$
    
    
    Therefore 
    
    \begin{equation} \label{eq:norm}
        \int_K a(y) \log \int_K a(x) Z(y, x) d x d y \le  d  \log (2 \pi e \eta^2 )
    \end{equation}
    
    Combining everything we arrive at 
    
    \begin{equation}
        d_{KL} (p, q) \le \frac {M^2} {2 \eta^2} + d \log \left ( \eta^3 (2 \pi)^{3/2} e^{1/2} \right )
    \end{equation}
    
    If $\eta \le (2 \pi)^{-1/2} e^{-1/6}$ then $d_{KL} (p, q) \le \frac {M^2} {2 \eta^2}$, since the second term is negative. 
    
    \emph{Case 2: Diagonal covariance matrices with eigenvalues in the range $[\eta_{\min}^2, \eta_{\max}^2]$}. The mixture models have expressions $p(x) = \int_K a(y) \mathcal N (x | \mu_p(y), \eta_p^2(y) I_d) dy$, and $q(x) = \int_K b(y) \mathcal N(x | \mu_q(y), \eta_q^2(y) I_d) dy$ where for all $y \in K$ we have that $\eta_p^2(y), \eta_q^2(y) \in [\eta_{\min}^2, \eta_{\max}^2]$. Moreover, let $\vartheta = (\eta_{\max} / \eta_{\min})^2 \ge 1$. We start by calculating the entropy term to be 
    
    \begin{equation}
        \int_K a(y) \ev {x \sim \nrv {\mu_p(y)} {\eta^2 I_d}} {- \log \nrv {x | \mu_p(y)} {\eta_p(y)^2 I_d}} d y \le \frac {d} 2 \log (2 \pi e \eta_{\max}^2)
    \end{equation}
    
    Similarly the terms in the numerator of the first term of Eq. 20 in \cite{durrieu2012lower} are bounded above by $d \log (2 \pi \eta_{\max}^2)$. Finally it remains to determine the denominator term. For that let $y_1, y_2 \in K$. We calculate the KL Divergence between $\nrv {\mu_p(y_1)} {\eta_p^2(y_1) I_d}$ and $\nrv {\mu_q(y_2)} {\eta_q^2(y_2) I_d}$ to be 
    
    \begin{equation}
        d_{KL} (\nrv {\mu_p(y_1)} {\eta_p^2(y_1) I_d}, \nrv {\mu_q(y_2)} {\eta_q^2(y_2) I_d}) \le \frac d 2 \left [ \vartheta - \log \vartheta - 1 \right ] + \frac {M^2} {2 {\eta_{\min}^2}} \le \frac {d \vartheta} 2  + \frac {M^2} {2 {\eta_{\min}^2}}
    \end{equation}
    
    Putting everything together we arrive at 
    
    \begin{equation}
        d_{KL} (p, q) \le \frac {M^2} {2 \eta_{\min}^2} + \frac d 2 \left [ \vartheta + 3 \log (2 \pi e \eta_{\max}^2) \right ]
    \end{equation}
    
    \emph{Application of the bound.} Finally, applying the above in the case that $\eta_{\min}^2 = \eta^2$ and $\eta_{\max}^2 \le (\ell + 1)^2 \eta^2$ we get that 
    
    \begin{equation}
        d_{KL} (p, q) \le \frac {M^2} {2 \eta^2} + \frac d 2 \left [ (\ell + 1)^2 + 3 \log (2 \pi e (\ell + 1)^2 \eta^2) \right ]
    \end{equation}
    
    The term $(\ell + 1)^2 + 3 \log (2 \pi e (\ell + 1)^2 \eta^2)$ becomes $\le 0$ when 
    
    \begin{equation}
        \eta \le \frac {\exp(-(\ell + 1)^2 / 6)} {(2 \pi e)^{1/2} (\ell + 1)}
    \end{equation}

\end{proof}

\subsubsection{Proof of Lemma \ref{lemma:px_py}} \label{sec:px_py}
    
\begin{proof}
    \textbf{At most one reflection occurs.} Let $\mathcal P_x$ and $\mathcal P_y$ denote the corresponding distributions. By Pinkser's inequality we have 
    \begin{equation}
    \begin{split}
        \| \mathcal P_x - \mathcal P_y \|_{TV} & \le \sqrt {\frac 1 2 d_{KL}(\mathcal P_x, \mathcal P_y)} \\
    \end{split}
    \end{equation}

    Recall that for two Gaussians $\nrv {\mu_x} {\Sigma_x}$ and $\nrv {\mu_y} {\Sigma_y}$ their KL divergence is equal to (supplementary Lemma \ref{lemma:kl}) 
    
    \begin{equation}
        d_{KL} \left ( \nrv {\mu_x} {\Sigma_x}, \nrv {\mu_y} {\Sigma_y} \right ) = \frac 1 2 \left [ \log \frac {|\Sigma_y|} {|\Sigma_x|} - d + \mathrm {tr}(\Sigma_x \Sigma_y^{-1}) + (\mu_y - \mu_x)^\top \Sigma_y^{-1} (\mu_y - \mu_x) \right ] 
    \end{equation}

    And for $\Sigma_x = \Sigma_y = \sigma^2 I_d$ the KL divergence becomes $\frac 1 {2 \sigma^2} \| \mu_x - \mu_y \|^2 $. We investigate multiple cases for what can happen to $x$ and $y$. 
 
    \emph{Case 1: $\tilde x, \tilde y \in K$ (no reflection).} First of all, the two proposal points $\tilde x$ and $\tilde y$ may not lie outside the convex body and hence the step occurs similarly to the un-truncated case. The probability of this event happening admits a Chernoff-type bound which depends on the sum of the distances of $\tilde x$ and $\tilde y$ from the boundary of the convex body $K$. We describe the first case, which yields results similar to the analysis of \cite{dwivedi2019log} and \cite{lee2020logsmooth}. The covariances are equal to $\eta^2 I_d$ and for the means we have that 
    
    \begin{equation}
        \begin{split}
            \| \mu_x - \mu_y \| & = \left \| x - y - \frac {\eta^2} 2 (\nabla f(x) - \nabla f(y)) \right \| \\
            & \le \| x - y \| + \frac {\eta^2 L} 2 \| x - y \| \\
            & \le \left ( 1 + \frac {\eta^2 L} 2 \right ) \eta
        \end{split}
    \end{equation}

    Where we have used the triangle inequality and smoothness. Plugging everything 
    
    \begin{equation}
            \| \mathcal P_x - \mathcal P_y \|_{TV} \le \frac {1 + \frac {\eta^2 L} {2}} {4}
    \end{equation}

    \emph{Case 2: Both points reflect.} This event happens where both the proposal points lie outside $K$ and points intersecting the boundary at two infinitesimal surfaces $d S_{\tilde x}$ and $d S_{\tilde y}$ with probability masses equal to the respective integrals over the boundary density $a$. An upper bound for the displacement of the means is twice the distance of Case 1, when the surfaces have anti-parallel normals and points reflect almost tangentially to the boundaries. The distance of the conditional densities is $\| \mu_x  - \mu_y \| \le \left ( 1 + \frac {\eta^2 L} 2 \right ) 2 \eta$. Note that one here can use the Pythagorean Theorem to improve the bound, however computations will become more complicated. 
    
    \emph{Case 3: One of the two points reflect.} The means' distance is at most the one of Case 2, hence the distance between the conditional densities is at most the one of Case 2. 
    
    \emph{Overall.} The minimum variance is $\eta^2 I_d$ (no reflection) and the maximum variance is $4 \eta^2 I_d$ (due to reflection). If $
        \eta \le \frac {\exp(-(\ell + 1)^2 / 6)} {(2 \pi e)^{1/2} (\ell + 1)}$, then from Lemma \ref{lemma:kl_mixtures} for $M = 2 \left ( 1 + \frac {\eta^2 L} 2 \right ) \eta$, we obtain that $\| \mathcal P_x  - \mathcal P_y \|_{TV} \le \frac 1 2 \left ( 1 + \frac {\eta^2 L} 2 \right ) \le \frac 1 2 \left ( 1 + \frac 1 {2c} \right )$.

    \textbf{Multiple Reflections.} For the means, in the worst case, each reflection adds a displacement of $\left ( 1 + \frac {\eta^2 L} 2 \right ) {\eta}$. Counting the zero-th step we get an upper bound of $\| \mu_x - \mu_y \| \le \left ( 1 + \frac {\eta^2 L} 2 \right ) \eta (\ell + 1)$. For the variances we denote with $\{ x_j \}_{j \in [\ell_x]}$ and $\{ y_j \}_{j \in [\ell_y]}$, for $\max \{ \ell_x, \ell_y \} \le \ell $ the sequences of the reflections, with $a_{xj} = x_j - x_{j - 1}$ and $a_{yj} = y_j - y_{j - 1}$  the corresponding rays, and with $\{ n_{xj} \}_{j \in [\ell_x]}$ and $\{ n_{yj} \}_{j \in [\ell_y]}$ the corresponding normals. The reflection operation obeys the following recurrence relation for $x_j$ (and for $y_j$ respectively): $a_{xj} = -2 (a_{x,j - 1}^\top n_{xj}) n_{xj} + a_{x,j - 1}$. Similarly to Lemma \ref{lemma:px_py} the covariance term cancels out with the first variance term (independently of the normal), hence $\var {} {a_{xj}} = \var {} {a_{x,j - 1}} = \dots = \var {} {a_{x,1}} = \eta^2 I_d$\footnote{Intuitively, the ray which is proportional to the velocity by a factor of $\eta$, undergoes a rotation, since we can think the reflection as applying a rotation operator to the velocity. The rotation does not change the diagonal covariance of the Gaussian.}. Using Cauchy-Swarchz we obtain an upper bound $\var {} {y_j} = \var {} {x_j} \preceq (1 + j)^2 \eta^2 I_d \preceq (1 + \ell)^2 \eta^2 I_d$. The bound agrees with the base case of $\ell = 1$. 
\end{proof}

\subsection{Proof of Lemma \ref{lemma:px_tx}} \label{sec:px_tx}

\begin{proof}
    For convenience define $G = \sqrt {L d} C \log (\kappa / \varepsilon)$ such that $\| \nabla f(x) \| \le G$ for all $x \in \Omega$. By Lemma \ref{lemma:tv_px_tx} the total variation distance obeys
    
    \begin{equation} \label{eq:px_tx_initial}
        \begin{split}
            \| \mathcal P_x - \mathcal T_x \|_{TV} & = 1 - \ev {v \sim \nrv 0 {I_d}} {\exp (\mathcal H(x, v) - \mathcal H(\tilde x', \tilde v'))} \\
        \end{split}
    \end{equation}
    
    Since $\exp (\mathcal H(x, v) - \mathcal H(\tilde x', \tilde v)) \ge 0$ from conditional expectation
    \begin{equation} \label{px_tx_next}
        \begin{split}
             \| \mathcal P_x - \mathcal T_x \|_{TV} & \le 1 - \ev {v \sim \nrv 0 {I_d}} {\exp (\mathcal H(x, v) - \mathcal H(\tilde x', \tilde v'))} \\
             & \le 1 - \pr {v \sim \nrv 0 {I_d}} {\| v \| \le V, \tilde x \in \Omega} \ev {v \sim \nrv 0 {I_d}} {\exp (\mathcal H(x, v) - \mathcal H(\tilde x', \tilde v')) | \| v \| \le V, \tilde x' \in \Omega}
        \end{split}
    \end{equation}

    \paragraph{Determine $\pr {v \sim \nrv 0 {I_d}} {\| v \| \le V, \tilde x \in \Omega}$.} In Lemma \ref{thm:chi_2_concentration} we choose $t = \sqrt {18 / d}  \log 2  < 3$ hence 
    
    \begin{equation}
        \Pr \left [ (1 - \sqrt {18/d} \log 2) d \le \| v \|^2 \le (1 + \sqrt  {18/d}  \log 2) d \right ] \ge 1 - \frac 2 {e^3} \ge \frac 9 {10} 
    \end{equation}
    
    Therefore with very high probability $\| v \|^2 \le 3d$. Conditioned on that we can get that the remaining probability is at least $1 - 3d^{-G} > 1 - (\kappa / \varepsilon)^{-Cd}$ due to Theorem \ref{thm:gradient_norm_concentration}. Now we are going to bound the change in the potential energy $f(x)$ and in the kinetic energy $\frac 1 2 \| v \|^2$, assuming that $V^2 < 3d$. 
    
    \paragraph{Bounding the change in $f(x)$.} The pairwise bounds between the distances of  $x, \tilde x, \tilde x'$ are
   $     \| \tilde x - x \|  = \| \eta \hat v \| = \eta \| \hat v \| = \eta \left \| v - \frac {\eta} 2 \nabla f(x) \right \| \le \eta \left ( \| v \| + \frac \eta 2 \| \nabla f(x) \| \right ) \le \eta V + \frac {\eta^2} 2 \| \nabla f(x) \| \le \eta V + \frac {\eta^2} 2 G $ and thus $\| \tilde x' - x \|  \le (k + 1) \left ( \eta V + \eta^2 G / 2 \right ) \le (\ell + 1) \left ( \eta V + \eta^2 G / 2 \right ) $ since by assumption $k \le \ell$. Thus we get
    
    \begin{align} \label{eq:u_change}
        f(\tilde x') - f(x) & \le   \nabla f^\top (x) ( \tilde x' - x )  + \frac L 2 \| \tilde x' - x \|^2 \tag*{(due to smoothness definition)} \\
        & \le  \| \nabla f(x) \| \| \tilde x' - x \| + \frac L 2 \| \tilde x' - x \|^2 \tag*{(Cauchy-Schwarz inequality)} \\
        & \le  (\ell + 1) G (\eta V + \eta^2 G / 2) + \frac L 2 (\ell + 1)^2 (\eta V + \eta^2 G / 2)^2 \tag*{(upper bounds derived above)} \\
        & = O(\tau) + \text{terms of the form } O \left ( \frac {\tau} {d^k \ell^r} \right )\tag*{}
    \end{align}
          
    where we set $\tau \sdef C / \sqrt c < 1$ for $c \ge 1$.

    \paragraph{Bounding the change in $\frac 1 2 \| v \|^2$.} We know that 
    
    \begin{align}
            \frac 1 2 \left \| \tilde v' \right \|^2 - \frac 1 2 \| v \|^2 & =  \frac 1 2 \left \| \hat v' - \frac \eta 2 \nabla f(\tilde x') \right \|^2 - \frac 1 2 \| v \|^2 \tag*{(Leapfrog integrator)} \\
            & \le \frac 1 2 \left ( \| \hat v' \| + \frac \eta 2 \left  \| \nabla f (\tilde x') \right \|  \right )^2 - \frac 1 2 \| v \|^2 \tag*{(Triangle inequality)} \\
            & = \frac 1 2 \left ( \| \hat v \| + \frac \eta 2 \left  \| \nabla f (\tilde x') \right \|  \right )^2 - \frac 1 2 \| v \|^2 \tag*{(Reflection preserves length)}\\
            & \le \frac 1 2 \| \hat v \|^2 - \frac 1 2 \| v \|^2 + \frac {\eta^2} 8 \| \nabla f(\tilde x') \|^2 + \frac \eta 2 \| \hat v \| \| \nabla f(\tilde x') \| \tag*{(expand $(x + y)^2$)} \\
            & \le \frac 1 2 \| \hat v \|^2 - \frac 1 2 \| v \|^2 + \frac {\eta^2 G^2} 8 + \frac \eta 2 G (V + \eta G) \tag*{(use fact $\| \nabla f(x) \| \le G$)} \\
            & \le \frac 1 2 \left \| v - \frac \eta 2 \nabla f(x) \right \|^2 - \frac 1 2 \| v \|^2 + \frac {\eta^2 G^2} 8 + \frac \eta 2 G (V + \eta G) \tag*{(Leapfrog integrator)}  \\
            & = \frac 1 2 \left [ \| v \| + \frac \eta 2 \| \nabla f(x) \| \right  ]^2 - \frac 1 2 \| v \|^2  + \frac {\eta^2 G^2} 8 + \frac \eta 2 G (V + \eta G) \tag*{(Triangle inequality)} \\
            & \le \frac \eta 2  V G + \frac {\eta^2 G^2} 8 + \frac {\eta^2 G^2} 8 + \frac \eta 2 G (V + \eta G) \tag*{(apply bounds)} \\
            & = \frac 3 4 \eta^2 G^2 + \eta V G \tag*{(Simplify)} \\
            & = \text{terms of the form } O \left ( \frac {\tau} {d^k \ell^r} \right ) \tag*{}
    \end{align}

    where we have used the fact that the reflection of the velocity preserves its norm.
    For small $\tau$ and for large value of $d$ and $\ell$ the Hamiltonian roughly behaves as $O(\tau)$.

    \begin{equation}
        \left \| \mathcal P_x - \mathcal T_x \right \|_{TV} \le 1 - \frac 9 {10} \exp (- O(\tau))
    \end{equation}
    
    So for small values of $\tau$, the above quantity is approximately $\tfrac 1 {10}$. 
    
\end{proof}

\section{Concentration Bounds}

We cite the following concentration bounds regarding $\chi^2$-variables and the behaviour of $\| \nabla f(x) \|$ where $x \sim \exp(-f(x))$ and $f$ is smooth and strongly convex. 

\begin{theorem}[Concentration of $\chi^2$ variables \cite{shalev2014understanding}] \label{thm:chi_2_concentration} Let $Z$ be a $\chi^2$-distributed random variable with $d$ degrees of freedom. Then for any $t > 0$ we have that $\Pr [Z \le (1 - t) d] \le \exp \left ( - t^2 d / 6 \right )$ and for $t \in (0, 3)$ we have that $\Pr [Z \ge (1 + t) d] \le \exp \left ( - t^2 d / 6 \right )$. Moreover for any $t \in (0, 3)$ we have that 

\begin{equation}
    \Pr [(1 - t) d \le Z \le (1 + t) d] \ge 1 - 2 \exp \left ( - t^2 d / 6 \right )
\end{equation}

\end{theorem}


\begin{theorem}[Gradient Norm Concentration \cite{lee2020logsmooth}] \label{thm:gradient_norm_concentration}
    
    Let $f: K \to \mathbb R$ be an $L$-smooth twice-diffferentiable function and $\pi$ be a density such that $\pi(x) \propto \exp(-f(x))$. Then for all $c > 0$ we have that 
    
    \begin{equation}
        \Pr_\pi \left [ \| \nabla f(x) \| \ge \ev {\pi} {\| \nabla f(x) \|} + c \sqrt {L} \log d \right] \le 3d^{-c}
    \end{equation}

    and subsequently

    \begin{equation}
        \Pr_\pi \left [ \| \nabla f(x) \| \ge \sqrt {Ld} + c \sqrt {L} \log d \right] \le 3d^{-c}
    \end{equation}

\end{theorem}


\section{Proof of Main Result (Theorem \ref{thm:main_result})} \label{sec:main_result}

\begin{proof}
    For the given $\eta$ we clearly have that $\eta < m^{-1/2}$ and for the set $\Omega$ we have concluded that for the given value of the constants $s < (\kappa / \varepsilon)^{-d}$. Moreover
    
    \begin{equation}
        \int_{c_0}^{1/4} \phi(x) d x = \int_{c_0}^{1/4} \frac {2^{16}} {a^2 \eta^2 m x \log (1 / x)} d x = \frac {2^{16}} {a^2 \eta^2 m} (\log \log (1 / c_0) - \log \log 4) 
    \end{equation}
    
    The function $\phi(t)$ attains a minimum at $t = 1/e$ with value and for $x \in [1/4, 1/2]$ we have that 
    
    \begin{equation}
        \phi(x) \le \phi(1/2) = \frac {2^{17}} {a \eta^2 m \log 2} = M 
    \end{equation}
    
    Combining everything into the bound we get that 
    
    \begin{equation}
        \begin{split}
            \| \nu_k - \pi \|_{TV} & \le \beta c_0 + \frac {2^{21}} {a^2 \eta^2 m k} \left [ \log \log (1 / c_0) - \log \log 4 + \frac {2} {\log 2} \right ] \\
            & \le \beta c_0 + \frac {2^{21} \cdot c \cdot \kappa d^2 (\ell + 1)^2 \log^2 (\kappa / \varepsilon)}  {a^2 k} \left [ \log \log (1 / c_0) - \log \log 4 + \frac {2} {\log 2} \right ]
        \end{split}
    \end{equation}
    
    Letting $c_0 \propto \frac {\varepsilon} {\beta}$ and $k$ a large multiple of $\kappa d^2 (\ell + 1)^{2} \log^2(\kappa / \varepsilon) \log \log (\beta / \varepsilon) \log(1 / \varepsilon)$ we can get that $\| \nu_k - \pi \|_{TV} \le \varepsilon$, which implies that $\| \pi_k - \pi \| \le \varepsilon$. The $\log (1 / \varepsilon)$ factor is used to boost the accuracy from $1 / (2e)$ to $\varepsilon$ as in~\cite{lee2020logsmooth}.  
    
\end{proof}

\section{Technical Lemmas}

Here we prove technical lemmas which we use in our analysis

\begin{lemma} \label{lemma:kl}
    Let $\nrv {\mu_x} {\Sigma_x}$ and $\nrv {\mu_y} {\Sigma_y}$ be two multivariate $d$-dimensional Gaussians. Then 
    
    \begin{equation}
        d_{KL} \left ( \nrv {\mu_x} {\Sigma_x}, \nrv {\mu_y} {\Sigma_y} \right ) = \frac 1 2 \left [ \log \frac {|\Sigma_y|} {|\Sigma_x|} - d + \mathrm {tr}(\Sigma_x \Sigma_y^{-1}) + (\mu_y - \mu_x)^\top \Sigma_y^{-1} (\mu_y - \mu_x) \right ] 
    \end{equation}

\end{lemma}

\begin{proof}
    The ratio of the densities is 
    
    \begin{equation}
        \left (\frac {|\Sigma_y|} {|\Sigma_x|} \right )^{1 / 2} \exp \left ( - \frac 1 2 (x - \mu_x)^\top \Sigma_x^{-1} (x - \mu_x) + \frac 1 2 (x - \mu_y)^\top \Sigma_y^{-1} (x - \mu_y) \right )
    \end{equation}
    
    Taking logarithms we get
    
    \begin{equation}
        \frac 1 2 \log \frac {|\Sigma_y|} {| \Sigma_x |} - \frac 1 2 (x - \mu_x)^\top \Sigma_x^{-1} (x - \mu_x) + \frac 1 2 (x - \mu_y)^\top \Sigma_y^{-1} (x - \mu_y)
    \end{equation}
    
    Taking expectations with respect to $\nrv {\mu_x} {\Sigma_x}$ we get 
    
    \begin{equation}
        \frac 1 2 \log \frac {|\Sigma_y|} {| \Sigma_x |} + \frac 1 2 (\mu_x - \mu_y)^\top \Sigma_y^{-1} (\mu_x - \mu_y) + \frac 1 2 \mathrm {tr} (\Sigma_y^{-1} \Sigma_x - I_d) 
    \end{equation}
    
    Rearranging terms we finally get 
    
    \begin{equation}
        d_{KL} \left ( \nrv {\mu_x} {\Sigma_x}, \nrv {\mu_y} {\Sigma_y} \right ) = \frac 1 2 \left [ \log \frac {|\Sigma_y|} {|\Sigma_x|} - d + \mathrm {tr}(\Sigma_x \Sigma_y^{-1}) + (\mu_y - \mu_x)^\top \Sigma_y^{-1} (\mu_y - \mu_x) \right ] 
    \end{equation}

\end{proof}

\begin{lemma} \label{lemma:tv_px_tx}

    Let $x_t$ be a Markov Chain positioned at $x$ with proposal and transition densities $p_x(\tilde x)$ and $t_x(\tilde x)$ and accept-reject probability equal to $\alpha_x(\tilde x)$ defined on a common space $\Omega$. Then the total variation distance between $p_x$ and $t_x$ is exactly $1 - \ev {\tilde x \sim p_x} {a_x(\tilde x)}$.
    
\end{lemma}

\begin{proof}
    The sampler remains in $x$ with a probability of $1 - \int_\Omega a_x(\tilde x) d \tilde x$. Moreover we also have $t_x(\tilde x) = a_x(\tilde x) p_x(\tilde x)$ for $\tilde x \neq x$. Therefore the total variation distance equals 
    \begin{equation}
    \begin{split}
        \| p_x - t_x \|_{TV} & = \frac 1 2 \left (\int_\Omega p_x(\tilde x) d \tilde x + 1 - \int_\Omega a_x(\tilde x) d \tilde x + \int_\Omega |p_x(\tilde x) - t_x(\tilde x)| d \tilde x \right ) \\ & \overset {a_x(\tilde x) \le 1} =  \frac 1 2 \left (  2 - 2 \int_\Omega a_x(\tilde x) p_x(\tilde x) d \tilde x \right ) \\
        & = \int_\Omega (1 - a_x(\tilde x)) p_x(\tilde x) d \tilde x \\
        & = \ev {\tilde x \sim p_x} {1 - a_x(\tilde x)} \\
        & = 1 - \ev {\tilde x \sim p_x} {a_x(\tilde x)}
    \end{split}
    \end{equation}
    
\end{proof}


    
    

\begin{lemma} \label{lemma:entropy_gaussian}
    Let $X \sim \nrv {\mu} {\Sigma}$ be a random variable. Then the entropy of $X$ is 
    
    \begin{equation}
        \frac 1 2 \log ((2 \pi e)^d |\Sigma|)
    \end{equation}

\end{lemma}

\begin{proof}
    The density of $X$ is 
    
    \begin{equation}
        \frac {1} {(2 \pi)^{d / 2} |\Sigma|^{1/2}} \exp \left ( - \frac {1} {2} (x - \mu)^\top \Sigma^{-1} (x - \mu) \right )
    \end{equation}
    
    By taking negative logs we get 

    \begin{equation}
        \frac {1} {2} (x - \mu)^\top \Sigma^{-1} (x - \mu) + \log ((2 \pi)^{d / 2} |\Sigma|^{1/2} )
    \end{equation}
    
    Finally taking expectation we arrive at $\frac 1 2 \log ((2 \pi e)^d |\Sigma|)$

\end{proof}

\section{Experiments Addendum}

\subsection{Polytopes}

We experiment with the following polytopes 

\begin{itemize}
    \item 100-Cube. The 100-dimensional cube $[-1, 1]^{100}$.
    \item 100-Simplex. The 100-dimensional simplex $\Delta_{100} = \left \{ x \in \mathbb R^{100} | \sum_{i = 1}^{100} x_i \le 1, x_i \ge 0 \right \}$.
    \item 10-Birkhoff. Its vertices correspond to the perfect matchings of $K_{10, 10}$. The Birkhoff polytope is the convex hull of the indicator vectors $\{ \mathbf 1_M | M \text { is a perfect matching of } K_{10, 10} \}$. 
    \item 10-Cross. The 10-dimensional unit ball $\{ x \in \mathbb R^{10} | \| x \|_1 \le 1 \}$. 
    \item 50-P-Simplex. The product $\Delta_{50} \times \Delta_{50}$.
    \item 100-S-Cube. A skinny cube of the form $[-100, 100] \times [-1, 1]^{99}$.
    \item e-coli. The core Escherichia coli metabolic model. 
    \item iAB-RBC-283. A proteomically derived knowledge-base of erythrocyte metabolism.
    \item iAT-PLT-636. Metabolic polytope regarding the human platelet. iAT-PLT-636, is reconstructed using 33 proteomic datasets and 354 literature references. The network contains enzymes mapping to 403 diseases and 231 FDA approved drugs.
    \item Recon1. Human (homo sapiens) metabolic network.
\end{itemize}

The metabolic polytopes were initially presented in the form $A_{eq} x = b_{eq}, \; l \le x \le u$. To convert the expression to the form $A x \le b$ we took the following steps

\begin{itemize}
    \item We calculate the kernel of $A_{eq}$, $W$ where each column of $W$ is a column vector $w$ such that $Aw = 0$.
    \item We calculate the shift vector $x_s$ to be the solution to the underdetermined system $A_{eq} x = b_{eq}$.  
    \item We calculate $A$ and $b$ as 
    \begin{equation*}
        A = \begin{pmatrix} I_d \\ - I_d \end{pmatrix} W, \qquad b = \begin{pmatrix} u \\ -l \end{pmatrix} - \begin{pmatrix} I_d \\ -I_d \end{pmatrix} x_s
    \end{equation*}

\end{itemize}

\subsection{Extra Marginal Plots}

\begin{figure}[h]
\centering
\subfigure[]{\includegraphics[width=0.15\textwidth, ext=.png]{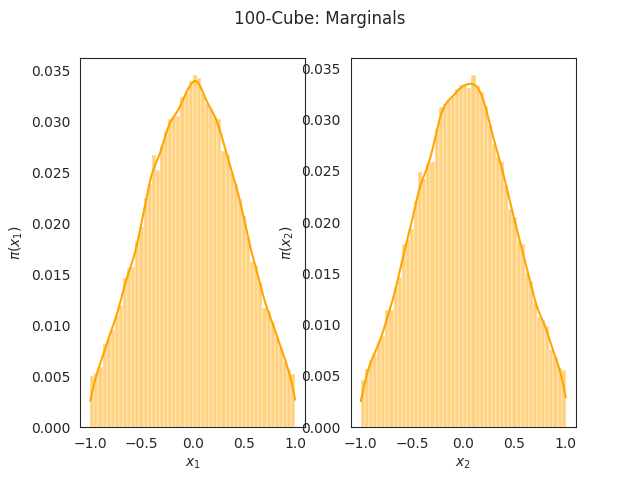}}
\subfigure[]{\includegraphics[width=0.15\textwidth, ext=.png]{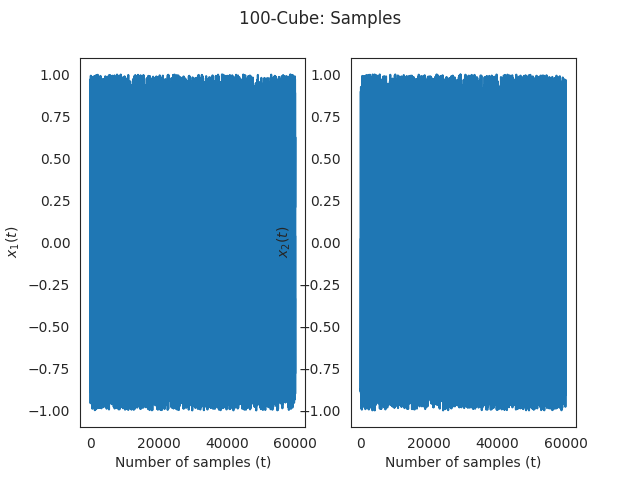}}
\subfigure[]{\includegraphics[width=0.15\textwidth, ext=.png]{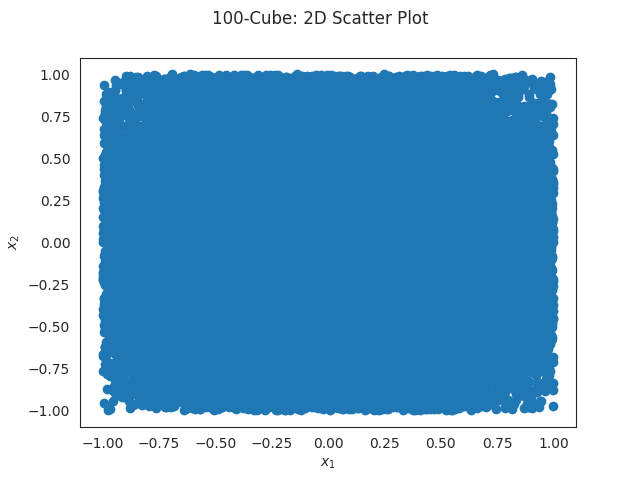}}
\subfigure[]{\includegraphics[width=0.15\textwidth, ext=.png]{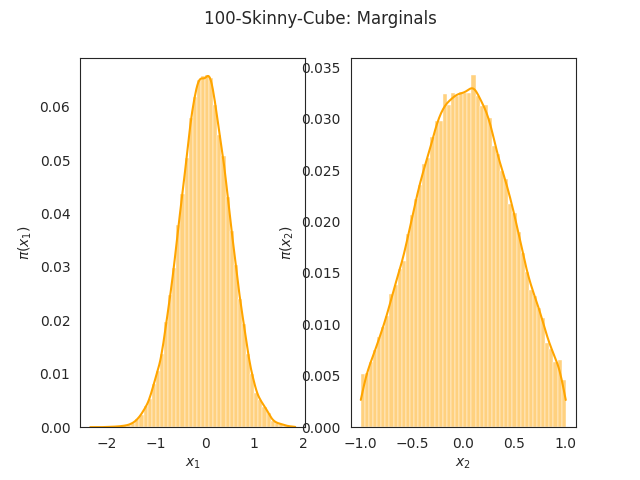}}
\subfigure[]{\includegraphics[width=0.15\textwidth, ext=.png]{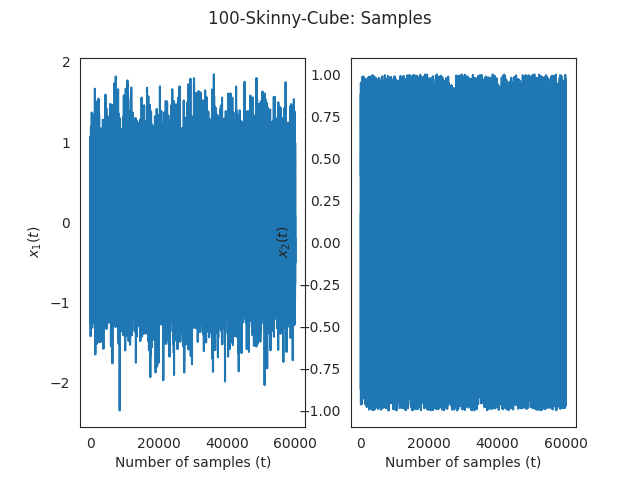}}
\subfigure[]{\includegraphics[width=0.15\textwidth, ext=.png]{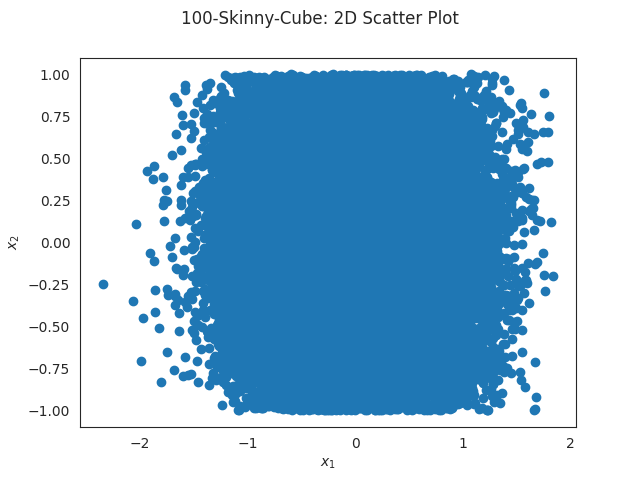}}
\subfigure[]{\includegraphics[width=0.15\textwidth, ext=.png]{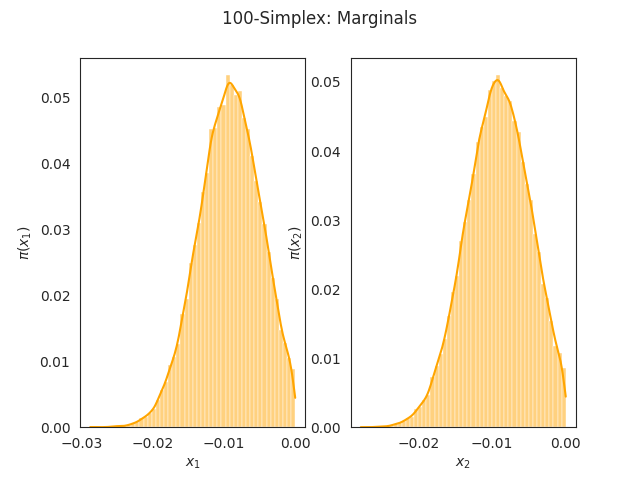}}
\subfigure[]{\includegraphics[width=0.15\textwidth, ext=.png]{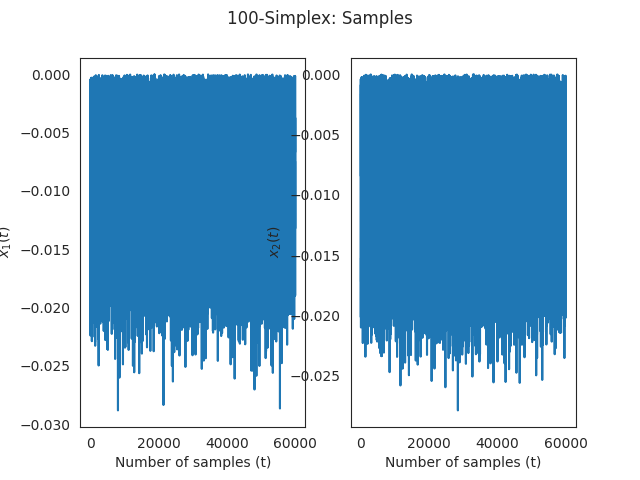}}
\subfigure[]{\includegraphics[width=0.15\textwidth, ext=.png]{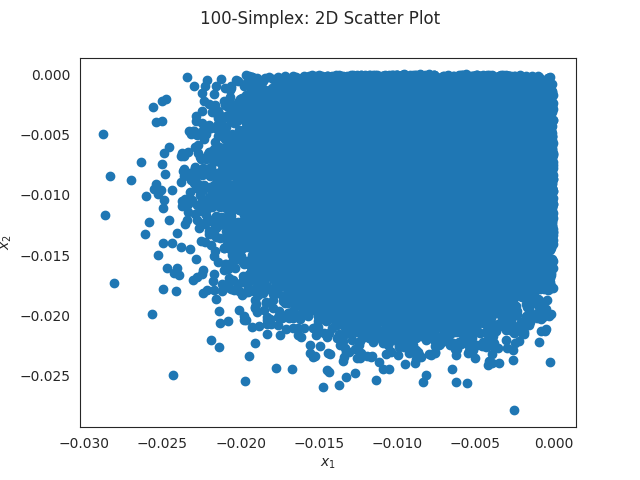}}
\subfigure[]{\includegraphics[width=0.15\textwidth, ext=.png]{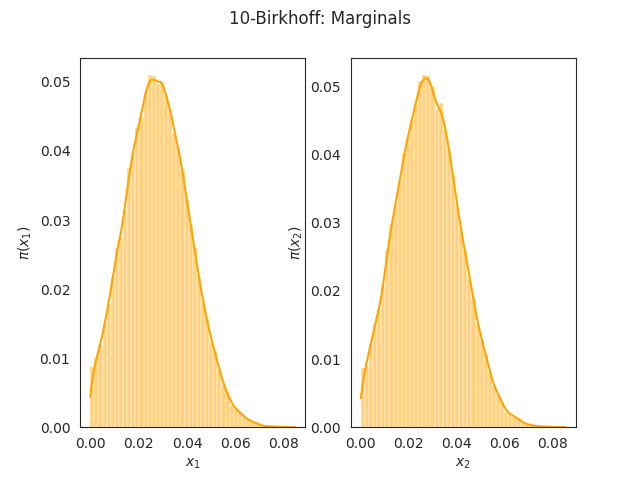}}
\subfigure[]{\includegraphics[width=0.15\textwidth, ext=.png]{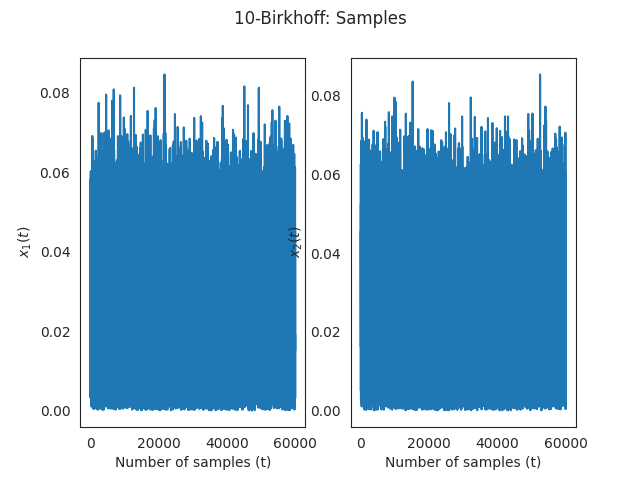}}
\subfigure[]{\includegraphics[width=0.15\textwidth, ext=.png]{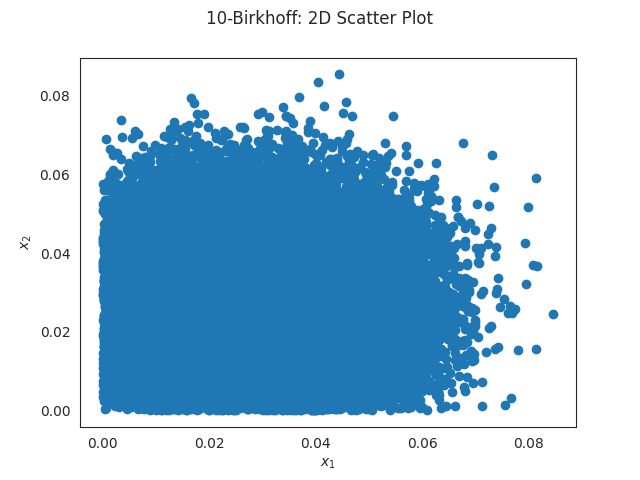}}
\subfigure[]{\includegraphics[width=0.15\textwidth, ext=.png]{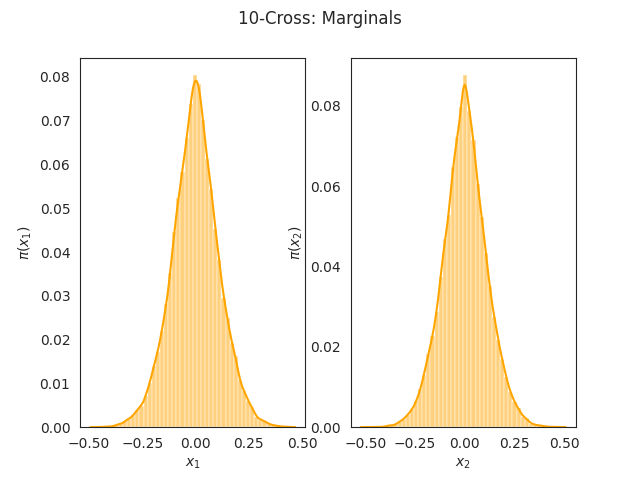}}
\subfigure[]{\includegraphics[width=0.15\textwidth, ext=.png]{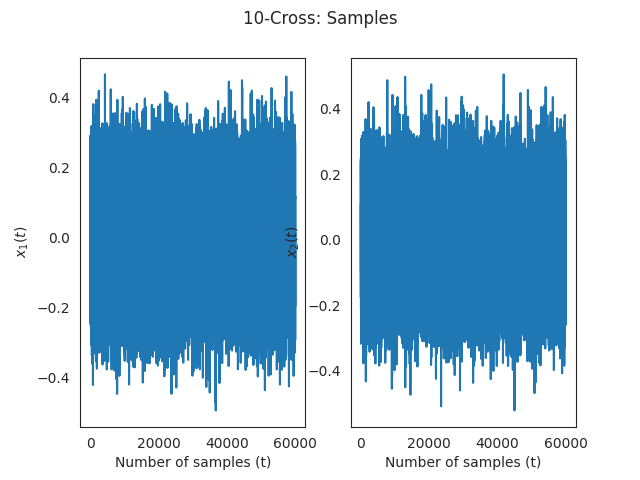}}
\subfigure[]{\includegraphics[width=0.15\textwidth, ext=.png]{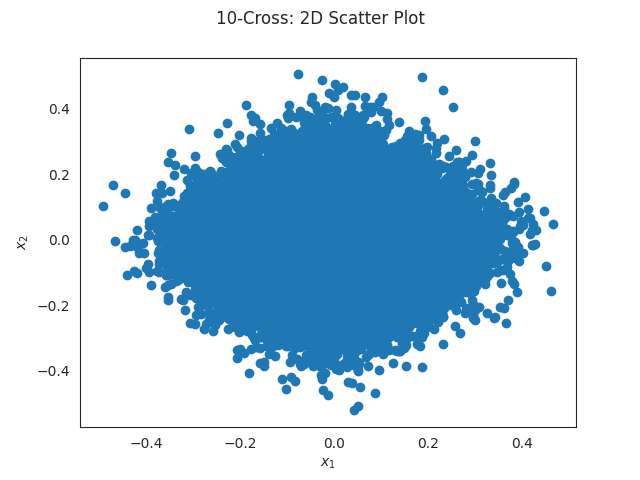}}
\subfigure[]{\includegraphics[width=0.15\textwidth, ext=.png]{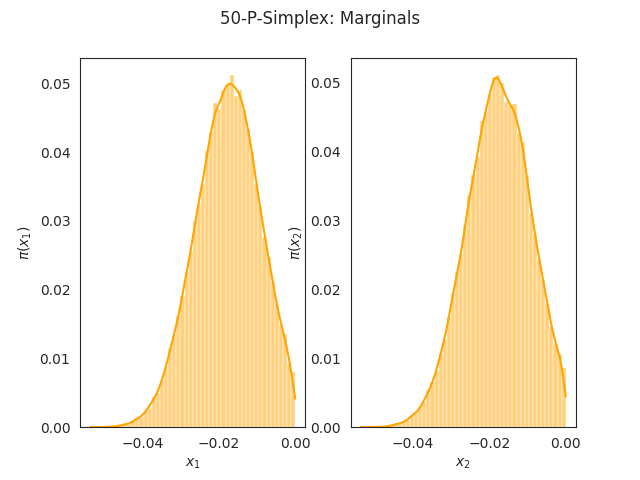}}
\subfigure[]{\includegraphics[width=0.15\textwidth, ext=.png]{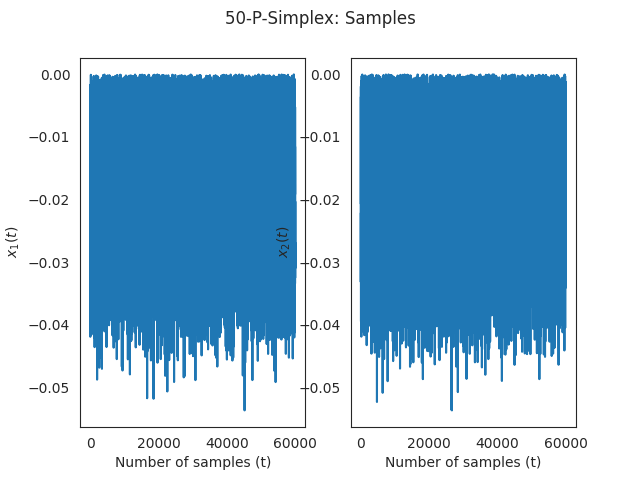}}
\subfigure[]{\includegraphics[width=0.15\textwidth, ext=.png]{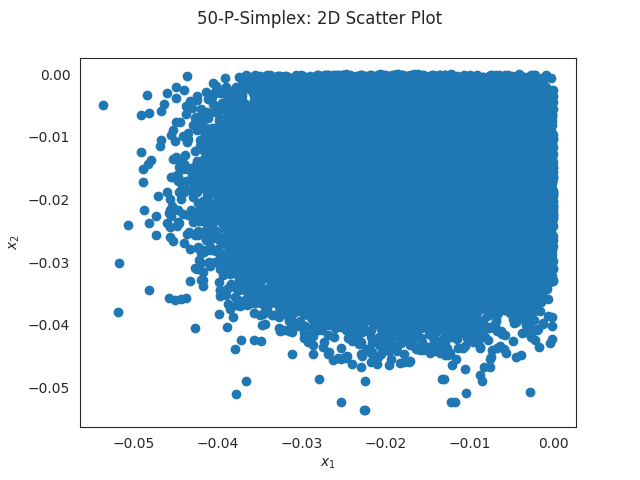}}

\caption{Marginal, scatter, and trace plots.}
\label{fig:marginals}
\end{figure}

In Figure \ref{fig:marginals} we provide marginal plots for the first two marginals ($x_1$ and $x_2$), trace plots, and 2D scatter plots for the density $\pi(x) \propto \exp \left ( -\tfrac {2 \| x - x_c \|^2} {R_c^2}\right )$. We have used an initial step size of $\eta_0 = R_c / 10$ and a walk length of $w = 100$.


\newpage


\end{document}

%% file: results.tex
\begin{tabular}{lrrrrrrrrr}
\toprule
                        & \rb{100-Cube}   & \rb{100-Simplex}   & \rb{100-S-Cube}   & \rb{10-Birkhoff}   & \rb{10-Cross}   & \rb{50-P-Simplex}   & \rb{e-coli}   & \rb{iAB-RBC-283}      & \rb{iAT-PLT-636}   \\
\midrule
$d$ & 100 & 100 & 100 & 81 & 10 & 100 & 25 & 130 & 290 \\
$\psi_{\mathrm{me}}$ & 1 & 1.0506 & 100 & 1.8 & 1 & 1.4 & 107.53 & $1.22 \cdot 10^6$ & 117.1559 \\
 \midrule 
                        &                 &                    &                   & ReHMC                &                 &                     &               &                       &                    \\
\midrule 
 $w$                    & 91            & 11               & 91              & 65               & 4             & 21                & 23          & 85                  & 225              \\
 $N_{\mathrm{ess}}$     & 24353         & 14666            & 8740            & 50309            & 50491         & 21076             & 15060       & 10                  & 9010             \\
 PSRF               & 1.001         & 1.004            & 1.001           & 1.002            & 1.001         & 1.003             & 1.001       & 1.102                & 1.002            \\
 $t_{\mathrm{is}}$ (us) & 1551         & \textbf{454}            & \textbf{5063}           & \textbf{396}            & \textbf{67}         & \textbf{502}             & \textbf{228}       & {$\mathbf{10.6 \cdot 10^6}$}           & \textbf{698320}           \\
 Avg. Num. Reflections ($\bar \ell$)          & 0.229       & 6.13            & 0.228          & 1.943            & 0.719        & 4.296              & 0.001    & 23.9943               & 0               \\
 Step size              & 0.008      & 0.001        & 0.008        & 0.004         & 0.058       & 0.0018          & 0.041     & 0.0003           & 0.003         \\
 \midrule 
                        &                 &                    &                   & H\&R-HOPS           &                 &                     &               &                       &                    \\
\midrule 
 $w$                    & 91           & 81              & 81             & 81              & 10           & 91               & 17         &   $\dagger$                    &    141                \\
 $N_{\mathrm{ess}}$     & 2799     & 164         & 629        & 665         & 16886    & 214          & 24     &    $\dagger$      &             467                   \\
 PSRF               & 1.003        & 1.020            & 1.006          & 1.006           & 1.001        & 1.016             & 1.178      & $\dagger$ & 1.011                               \\
 $t_{\mathrm{is}}$ (us) & 1608     & 20632       & 6315       & 3496        & 1301     & 16046        & 310060 &    $\dagger$                   &   $16 \cdot 10^6$            \\
 \midrule
                        &                 &                    &                   & CH\&R-HOPS          &                 &                     &               &                       &                    \\
                        \midrule 
 $w$                    & 91           & 91              & 91             & 81              & 10           & 91               & 17         &    $\dagger$                   &  197                  \\
 $N_{\mathrm{ess}}$     & 3225     & 293        & 71        & 4744        & 31734    & 17          & 11     &  $\dagger$                     & 453                   \\
 PSRF               & 1.002        & 1.022           & 1.016          & 1.001            & 1.000        & $(\star)$ 1.520             & 1.102      &  $\dagger$                     &   1.005                 \\
 $t_{\mathrm{is}}$ (us) & \textbf{396}      & 4061         & 20645      & 10609       & 351      & 194182        & 284494 &   $\dagger$                    &  $7.24 \cdot 10^6$                  \\
\bottomrule
\end{tabular}